\documentclass[11pt,a4paper]{article}
\usepackage{times,latexsym}
\usepackage{url}
\usepackage[T1]{fontenc}

\usepackage[acceptedWithA]{tacl2018v2}
\usepackage{xcolor}
\usepackage{graphicx}

\usepackage[utf8]{inputenc}
\usepackage[T1]{fontenc}
\usepackage[english]{babel}
\usepackage{algorithmic}
\usepackage[ruled,vlined,linesnumbered]{algorithm2e}
\usepackage{todonotes}
\usepackage{amsmath}
\usepackage{multirow}
\usepackage{booktabs}
\usepackage{tablefootnote}
\usepackage{setspace}
\usepackage{tabularx}
\usepackage{longtable}
\usepackage{fancyhdr}
\usepackage{pdflscape}
\usepackage{colortbl}
\usepackage{lipsum}
\usepackage{totcount}
\usepackage{subcaption}
\usepackage{float}

\usepackage{tikz}
\def\checkmark{\tikz\fill[scale=0.4](0,.35) -- (.25,0) -- (1,.7) -- (.25,.15) -- cycle;} 

\newtotcounter{citnum} 
\def\oldbibitem{} \let\oldbibitem=\bibitem
\def\bibitem{\stepcounter{citnum}\oldbibitem}

\usepackage{xspace,mfirstuc,tabulary}

\newif\iftaclinstructions
\taclinstructionsfalse 
\iftaclinstructions
\renewcommand{\confidential}{}
\renewcommand{\anonsubtext}{(No author info supplied here, for consistency with
TACL-submission anonymization requirements)}
\newcommand{\instr}
\fi

\iftaclpubformat 

\else

\fi

\title{A Survey on Artificial Intelligence for Source Code: A Dialogue Systems Perspective}

\author{Erfan Al-Hossami 
  \and Samira Shaikh\\
 Department of Computer Science \\
 University of North Carolina at Charlotte \\
 Charlotte, NC 28223 \\
  \texttt{\{ealhossa, sshaikh2\}@uncc.edu} \\
}

\date{}

\begin{document}
\maketitle
\begin{abstract}
In this survey paper, we overview major deep learning methods used in Natural Language Processing (NLP) and source code over the last 35 years. Next, we present a survey of the applications of Artificial Intelligence (AI) for source code, also known as Code Intelligence (CI) and Programming Language Processing (PLP). We survey over \total{citnum}\ publications and present a software-engineering centered taxonomy for CI placing each of the works into one category describing how it best assists the software development cycle. Then, we overview the field of conversational assistants and their applications in software engineering and education. Lastly, we highlight research opportunities at the intersection of AI for code and conversational assistants and provide future directions for researching conversational assistants with CI capabilities.
\end{abstract}


\section{Introduction \& Motivation}
\label{sec:intro}


Conversational Assistants, also known as task-oriented dialogue systems, are very widely used and accessible such as Siri and Alexa. These assistants have been increasingly used to assist human users in a variety of tasks such as reserving hotels, booking flights, or forecasting the weather. In recent years, we have also seen advancements in the field of Artificial Intelligence (AI) applied to source code also known as Code Intelligence (CI) and Programming Language Processing (PLP). Github Copilot\footnote{\href{https://copilot.github.com/}{https://copilot.github.com/}} powered by GPT-Codex~\cite{chen2021evaluating} is currently in beta as an Integrated Development Environment plugin that is able to assist software developers as a pair-programmer by suggesting code snippets or writing source code on its own. 
Given these advancements, perhaps task-oriented bots can be equipped with more capabilities to assist humans in cognitively demanding tasks, including programming by professionals or novices. By building models and tools that can generate both language and code, we could potentially better understand the cognitive basis of programming which can have key impacts on computer science education practices~\cite{fedorenko2019language}. This survey is structured as follows: Section \S\ref{sec:dl} explores general deep learning techniques that have been used to model language and source code over the last 35 years. Section \S\ref{sec:plp} surveys the field of CI with a systemic review (a) on datasets containing natural language and executable code (\S\ref{subsubsec:code_gen_datasets}), and (b) of all methods used to generate source code from natural language on the CoNaLa dataset, a popular python code generation benchmark (\S\ref{subsubsec:code_gen_methods}). Section \S\ref{sec:conv_ai} overviews the field of conversational artificial intelligence and its applications in software engineering and education. Finally, Section \S\ref{sec:future} highlights research opportunities at the intersection of CI and conversational assistants to provide future directions for research in this new area.

\section{Deep Learning Methods}
\label{sec:dl}

This section overviews major developments over the last 35 years in deep learning neural architectures that are used to generate and understand both natural language and source code. These deep learning methods can be used to generate natural language (e.g. by chatbots~\cite{zhang2019dialogpt}). In the context of code generation, deep learning methods have been used to generate code snippets (e.g. ~\cite{liguori2021evil,liguori2021shellcode,hijax2021,yin_tranx_2018}). The need for neural architectures arose as deep learning was tasked to solve specific problems such as machine translation, question-answering, and code generation. Neural architectures describe the general structure of the neural network(s), including how many layers it has and how units in these layers are connected to each other~\cite{sarkar_hands-transfer_2019}. Neural models can solve problems specific to the target domain through abstract modeling. For instance, in natural language, often, word order matters to the semantics of a sentence~\cite{payne1992pragmatics}. Furthermore, word order differences between languages can be problematic when translating from one language to another~\cite{Jurafsky2000SpeechAL}. How can neural architectures consider the order of the input sequence? We explore relevant neural architectures further in the following subsections.

\subsection{Multilayered Perceptrons}
\label{sec:mlp}

Here we introduce the multi-layered perceptron (MLP) as well as common terms associated with neural networks, including activation functions, loss functions, and gradient descent, and backpropagation.

\begin{figure}[ht!]
  \includegraphics[height=6cm,keepaspectratio]{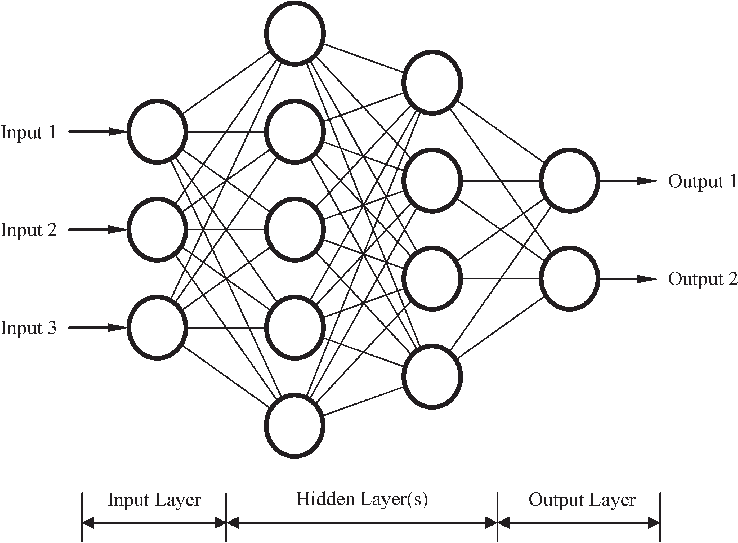}
  \caption{The Multilayer Perceptron architecture is composed of an input layer, an output layer and multiple hidden layers. Figure from~\cite{yuen2013intelligence}.}
  \label{fig:mlp}
\end{figure}

A multi-layered perceptron (MLP), commonly known as an artificial neural network, is a variant of the Perceptron model~\cite{rosenblatt_principles_1961}, composed of an input layer, multiple hidden layers fully connected with each other, and an output layer (c.f. Fig.~\ref{fig:mlp}). MLPs excel at learning the input representations and mapping them to the output variable. The input given to neural networks is numerical vectors. Often, categorical inputs are one-hot encoded.

Each layer is made up of several neurons. A neuron is a unit that takes in weighted inputs, a bias (an input of 1.0 with a computed weight), and computes an output using an activation function. Weights can be initialized to a constant number, a random number, or through more nuanced means such as the initialization proposed by He et al.~\citeyearpar{he2015delving}. Each neuron computes the dot product of its input vector $Z=(z_1,z_2,...z_n)$  and weight vector $W=(w_1,w_2,w_n)$ through $ x= Z\cdot W$ . Once the dot product is computed it is passed to an activation function (see table~\ref{tab:activations}) before being passed on as the output of the neuron. The process of neurons passing values forward in the network using their activation functions until an output is produced is called \textit{forward passing}.
\begin{table*}[ht!]
\centering
\begin{tabular}{|l|c|}
\hline
Activation Function & Definition \\ \hline
Binary Threshold    &   $f(x) =\left\{\begin{matrix}1\;  ,if\;   x>0\\ 
\;  0\; ,otherwise
\end{matrix}\right.$        \\ \hline
Sigmoid             &      $f(x) = \frac{1}{1+e^{-x}}$     \\ \hline
Tanh                &     $f(x) = \frac{e^{x} - e^{-x}}{e^{x}+e^{-x}}$      \\ \hline
ReLu                &     $f(x) =\left\{\begin{matrix}x\;  ,if\;   x>0\\ 
\;  0\; ,otherwise\;  OR \;  max\{0,x\}
\end{matrix}\right.$\\ \hline
\end{tabular}
\caption{Example activation functions used in neural networks}
\label{tab:activations}
\end{table*}
Lastly, another activation function called \textit{softmax} is commonly used in the \textit{output layer}. Given a vector $z$ from the last hidden layer, the  \textit{softmax} function produces a vector $\sigma(z)$ with length $K$. The numbers in $\sigma(z)$ are values between 0 and 1. Furthermore, the sum of all elements in $\sigma(z)$ equal to 1, $\sum_{j=1}^{K}\sigma(z)  = 1$. This property of softmax is often useful in classification tasks. The \textit{softmax} equation is illustrated below:
\begin{equation}
\label{eq:softmax}
\sigma(z_j) = \frac{e^{z_j}}{\sum_k e^{z_k}}
\end{equation} 

where $j = 1,2,...,K$. 

Once the MLP network produces an output (prediction), it is compared with the target values in the dataset (ground truth). This comparison yields a measurement of how well the output matches the expected values. There are several methods, called \textit{loss functions}, to measure how well the predicted output minimizes the error. Here are some examples of loss functions:
\begin{itemize}
\item \textbf{Binary cross-entropy:} It is also called the log-loss applied for a two class classification problem.
\item \textbf{Categorical cross-entropy:} log-loss applied for a classification task of N classes
\item \textbf{Mean Squared Error:} Mean of the squared sum of error, often used in regression tasks.
\item \textbf{Mean Absolute Error:} Average magnitude of the errors given a set of predictions. It gives higher weights to larger errors since it squares the errors.
\end{itemize}


\begin{figure}[ht!]
 \centering \includegraphics[width=\linewidth,height=7cm,keepaspectratio]{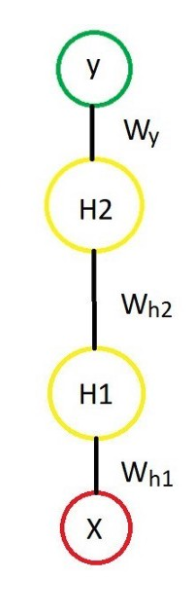}
  \caption{The MLP network has different and independent weights at each layer. $W_{h1}$ and $W_{h2}$ are different. Source: \href{https://medium.com/deep-math-machine-learning-ai/chapter-10-deepnlp-recurrent-neural-networks-with-math-c4a6846a50a2}{Medium post authored by Mady}.}
  \label{fig:feedforward} 
\end{figure}

In the MLP, the computed error is propagated back through each of the layers. This is done so that the weights for each neuron can be updated empirically. This process is called back-propagation~\cite{rumelhart_learning_1986}. The weights can be updated after each training example, in a process called \textit{online learning}. This process can result in big changes in the network from example to example and is largely unstable. Alternatively, computed errors can be saved for a set of training examples and the MLP network weights are updated after training on the set. This is called \textit{batch learning}. The learning rate $\alpha$ controls the size of a change made to a weight, so the weights do not change dramatically and learning can be more stable. 

MLP networks are trained using a training algorithm such as stochastic gradient descent (SGD)~\cite{bottou_large-scale_2010}. 

SGD is a simplified version of gradient descent that computes the gradient descent for only a small sample of training data to estimate the gradient descent given a loss function for the entire training dataset. These optimization algorithms are responsible for updating weights in a model given a training set sample and a loss function. Other algorithms are commonly used to further optimize SGD such as Adaptive Moment Estimation (Adam)~\cite{kingma_adam_2014}. Adam further optimizes SGD by computing adaptive learning rates for each parameter.
MLPs excel at mapping inputs to outputs, however, these inputs are assumed to be independent of one another. What if the inputs are dependent? What if their order matters such as in word sequences forming sentences? Next, we introduce another neural network architecture designed specifically to address this problem, the Recurrent Neural Networks.

\subsection{Recurrent Neural Networks}
\label{sec:rnn}

\begin{figure*}[ht!]
  \includegraphics[width=\linewidth]{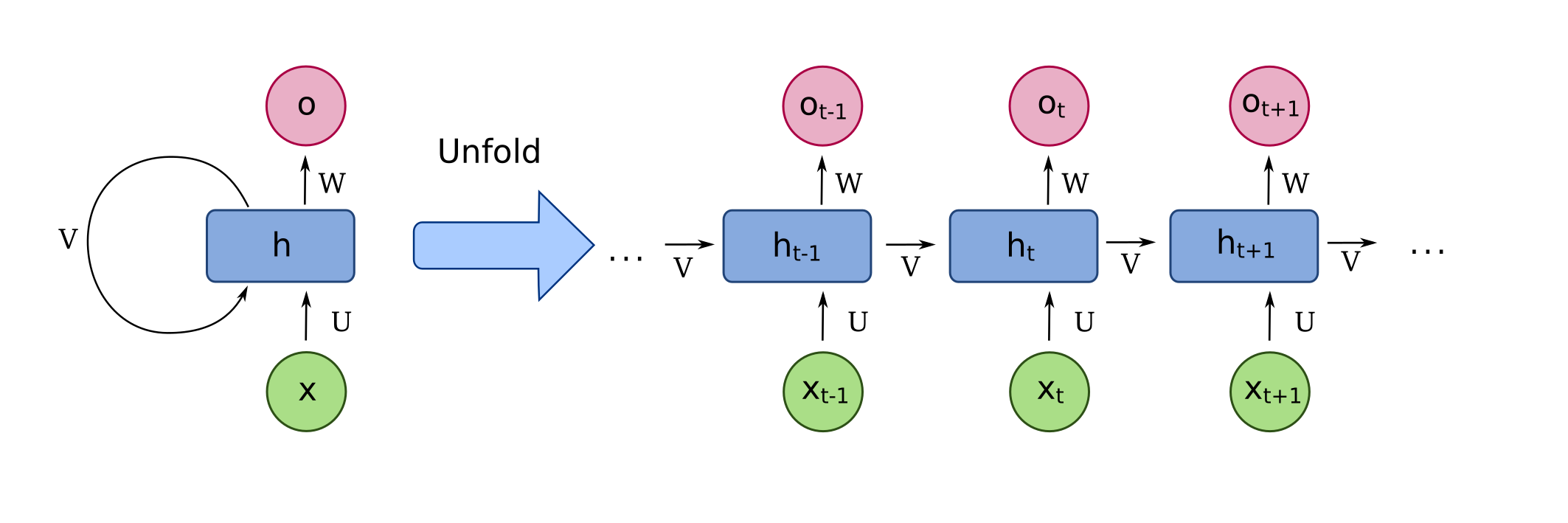}
  \caption{The Recurrent Neural Network architecture is composed of an input layer $x$, an output layer $o$, a hidden layer $h$, and a feedback loop $v$ for each time step $t$. The weights $v,u,w$ are not different for each layer. The weights are instead shared across all the time steps. Source: \href{https://upload.wikimedia.org/wikipedia/commons/thumb/b/b5/Recurrent_neural_network_unfold.svg/220px-Recurrent_neural_network_unfold.svg.png}{Wikipedia}.}
  \label{fig:rnn} 
\end{figure*} 

Recurrent Neural Networks (RNNs) are specialized network architectures for processing a sequence of input values that are dependent upon one another. RNNs are often applied in predicting the next word/token in a sequence of words, or in translating from one language to another. RNNs follow an architecture distinct from that of MLPs. RNNs are equipped with a \textit{feedback loop}. This feedback loop enables the RNN to share weights $v,u,w$ across different time steps, meanwhile, MLPs use a different set of parameters and weights each at each hidden layer. This is illustrated in Figure~\ref{fig:feedforward} where $W_{h1}$ and $W_{h2}$ are different. We can imagine that the RNN is performing the same computations in each time step, but with different inputs and updated weights. This setup greatly reduces the number of parameters required to learn a task. Figure~\ref{fig:rnn} showcases the RNN architecture with a single hidden unit. When unfolding the RNN, we are left with multiple feedforward MLP networks, a network at each time step $t$. The total number of time steps is equivalent to the length of the input sequence.

While MLPs compute the hidden layer values from the input values and weights exclusively, the hidden layers in RNNs are computed from the input value $x_t$ at time step $t$, weights, and the previous hidden layer value. This is illustrated below:
\begin{equation}
\label{eq:rnn}
\begin{matrix}
h_t  = \sigma(u*x_t+v*h_{t-1})\\
o_t = softmax(w*h_t)
\end{matrix}
\end{equation}
where $v$ is a weight vector for the different time steps, $\sigma$ as an activation function as described in Table~\ref{tab:activations}, $x_t$ is an input value at time step $t$, $u$ is a weight vector for the hidden layer, $w$ is a weight vector for the output layer, and $o_t$ is the output value at time $t$. Softmax is described in Equation~\ref{eq:softmax} and it is used on the product of the hidden state $h_t$ and the weight vector $w$. The error is computed using a loss function $E$ (loss functions are described in section~\ref{sec:mlp}) for output $o_t$ and target sequence $s_t$. This yields an error $E_t$. As for the total error across the different timestamps it is computed through summing all of $E_t$ for an input length of $n$ $\sum_{t=0}^{n}E_t$. The weights are updated using back-propagation~\cite{rumelhart_learning_1986} as described in section~\ref{sec:mlp}, with a small tweak, the current time step is dependent on the previous time step, so the back-propagation traverses back to the first time step. u,v, and w are updated using SGD~\cite{bottou_large-scale_2010}. There is a big drawback for RNNs however, as we go back to adjust the weights during back-propagation, the signal gets close to zero this is called \textit{vanishing gradient}, or it grows exponentially, this is called \textit{exploding gradient}. This is an issue in particular when we expect RNNs to 'remember' and keep track of long-term dependencies. For that researchers have developed a special type of RNNs, the Long-Short Term Memory (LSTM).

\subsection{Long-Short Term Memory}
\label{sec:lstm}
\begin{figure*}[ht!]
  \includegraphics[width=\linewidth]{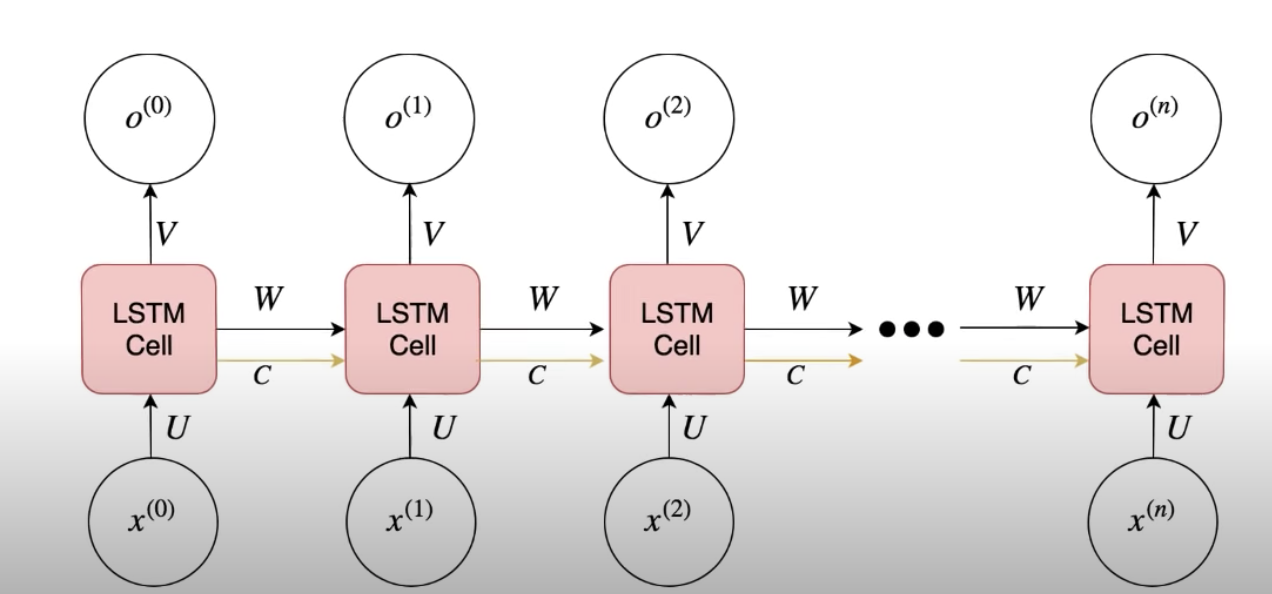}
  \caption{The Long-Short Term Memory (LSTM) network is similar to the Recurrent Neural Network (RNN) except the hidden state is replaced with an LSTM cell. Similarly to Figure~\ref{fig:rnn} $n$ represents a timestep, $x^{(n)}$ represents the input at time $n$, $U$ represents a weight vector for the LSTM cell, $W$ represents the weight vector in the feedback loop, $C$ represents the LSTM cell state, $V$ represents the weight vector for the output layer, and $o^{(n)}$ represents the output at time $n$. Source: \href{https://www.youtube.com/watch?v=QciIcRxJvsM}{CodeEmporium on Youtube}.}
  \label{fig:lstm_over} 
\end{figure*}

Recurrent Neural Networks (RNNs) lose historical context and dependencies over longer periods. To address the problem of memory in RNNs Hochreiter et al.~\citeyearpar{hochreiter_long_1997} proposed a novel RNN architecture called the Long-Short Term Memory (LSTM) in 1997. It excels at maintaining long-term dependencies and preventing problems such as the \textit{exploding} and \textit{vanishing gradient} issues. LSTMs are a variant, or a type, of RNNs. Their similarity is denoted in Figure~\ref{fig:lstm_over}.

\begin{figure}[ht!]
  \includegraphics[height = 3.75cm]{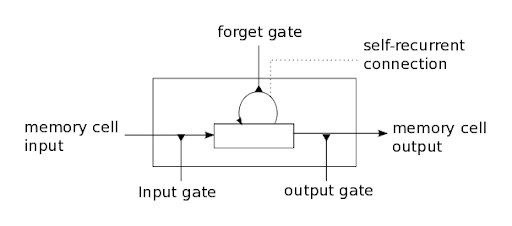}
  \caption{The LSTM is composed of 3 different gates: 1) an input gate controlling what is updated in the cell state, 2) a forget gate controlling what is omitted from the previous cell states (memory), and 3) an output gate controls what is passed on as the final output of the cell.}
  \label{fig:lstm_gates} 
\end{figure}

LSTMs contain \textit{gates} which enable the LSTM to add or remove information to a cell state. Each gate is like a one-layered neural network with each with its weight vector and biases. These networks then learn what information is relevant and what is not throughout training. LSTMs consist of three \textit{gates}: The input, output, and forget gates. As shown in Figure~\ref{fig:lstm_gates}, an LSTM cell is composed of 3 different gates. At a given time step, the input gate allows or denies incoming signals from the input layer. The output gate controls what gets passed on to other cells or layers as the cell's final output. The forget gate administers the self-recurrent connection to forget or remember previous cell states as deemed fit.

\begin{figure*}[ht!]
\centering
  \includegraphics[width=\linewidth,keepaspectratio]{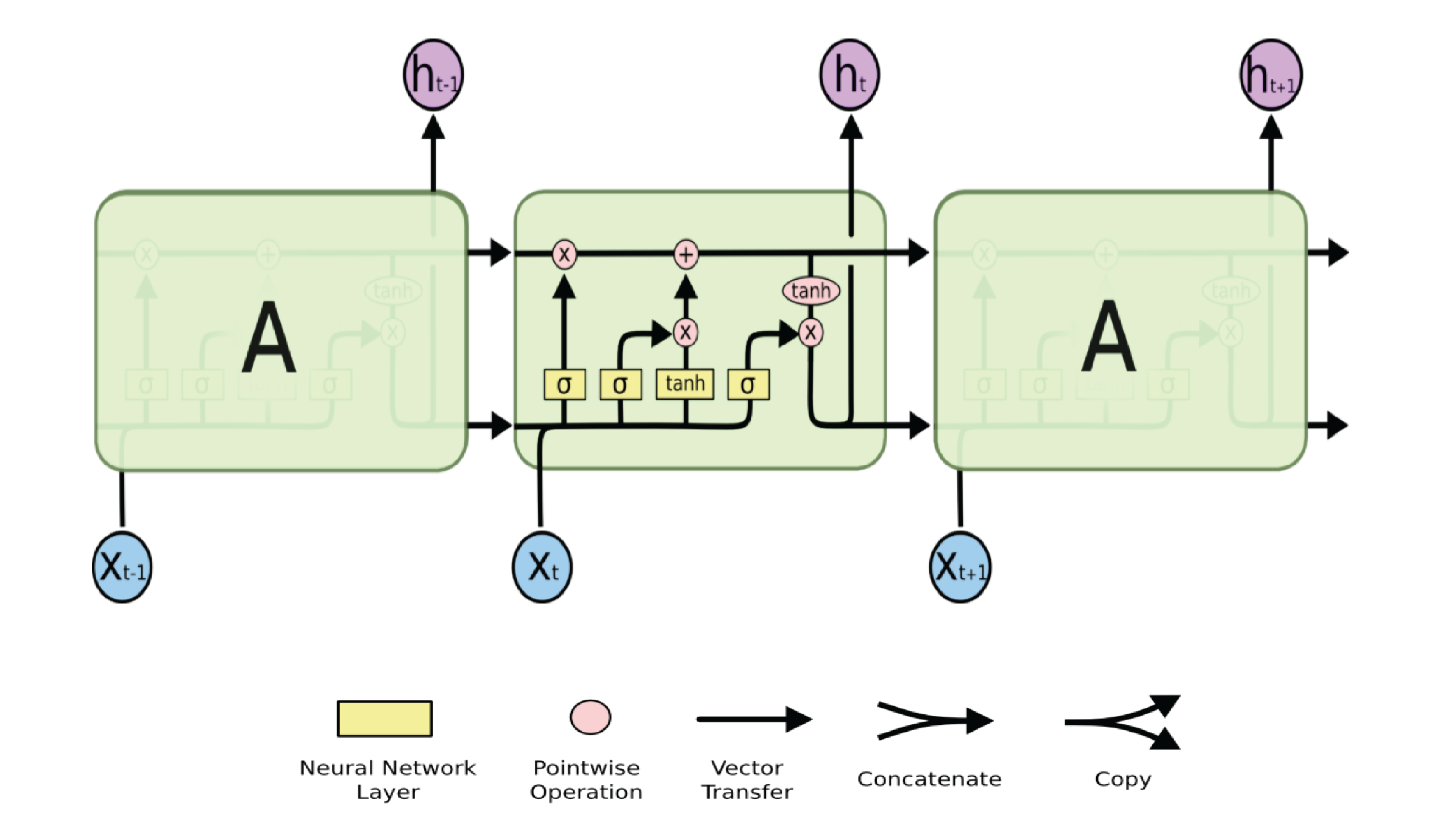}
  \caption{The Long-Short Term Memory (LSTM) cell architecture. At teach time step $t$, an LSTM cell takes in the previous cell state and the previous hidden state as inputs.  It contains multiple sigmoid and tanh functions to compute on those inputs for each of the 3 gates (input, forget, and output). The LSTM cell outputs, a cell state $C_t$, and a  hidden state $h_t$ that is passed onto other layers to produce an output $o_t$, and to LSTM cell's future self at time step $t+1$. Source: \href{http://colah.github.io/posts/2015-08-Understanding-LSTMs/}{Colah's blog}.}
  \label{fig:lstm_cell} 
\end{figure*} 

The LSTM cell architecture is displayed in Figure~\ref{fig:lstm_cell}. Let $C$ indicate a cell state, $h$ the hidden state for each time step $t$. Let $I$, $F$, and $O$, represent the input, forget, and output gates respectively. All gates are regulated by a sigmoid $\sigma$ function, which as mentioned earlier outputs a number between 0 and 1, controlling how much of the outputs of each gate should get passed along to the cell state. The processes of LSTM computations can be illustrated in four steps:
\begin{enumerate}
\item  \textbf{Deciding what to \textit{forget} and what to remember}: This decision is made by the forget gate at each time step. The forget gate utilizes information from the previous hidden state $h_{t-1}$ , the current input $x_t$, its own weight vector $W_F$, and its own bias $b_F$.  $h_{t-1}$ and $x_t$ are concatenated and they are multiplied in a dot product with $W_F$ and summed with $b_F$ . The result of that is then squashed through the sigmoid function $\sigma$. The result is $F_t$ a vector of numbers between 0 and 1, corresponding to each number in the previous hidden state and current input vectors, indicating how much should be remembered (if the value is greater than 0) and forgotten (if the value is 0). The process is described in mathematical notation in equation~\ref{eq:forget}.
\item  \textbf{Deciding what to write to memory}: This decision is handled by the input layer. The input layer utilizes information from the previous hidden state $h_{t-1}$ , the current input $x_t$, its own weight vector $W_I$, and its own bias $b_I$.  $h_{t-1}$ and $x_t$ are concatenated and then are multiplied in a dot product with $W_I$ and summed with $b_I$ . The result of that is then squashed through the sigmoid function $\sigma$. The result is $I_t$ a vector of numbers between 0 and 1, corresponding to each number in the previous hidden state and current input vectors, indicating \textbf{where} should new information should be \textbf{written} to the current cell state (memory). Next, another layer $\tilde{C_t}$, decides \textbf{what} new information are good candidates to be written to the current cell state using tanh as its activation function. The process is described in mathematical notation in equation~\ref{eq:input}.
\item \textbf{Updating the memory}: The current cell state is decided by utilizing the forget gate vector output $F_t$, erasing things that are determined to be forgettable from the previous cell state $C_{t-1}$, and the new information from step 2 is added after scaling by the input gate vector $I_t$.  The process is described in mathematical notation in equation~\ref{eq:cell}.
\item  \textbf{Output}: The final output for the LSTM cell is decided by the current input $x_t$, and the current cell state $C_t$. The output gate vector $O_t$ decides which parts of the cell state are going to be outputted and is computed using the previous hidden state $h_{t-1}$ and the current input $x_t$. Then, the current cell state $C_t$ is passed through a tanh function and multiplied by the output gate vector giving us $h_t$ which is passed on as the output of the LSTM to other cells and a copy to itself. The process is described in mathematical notation in equation~\ref{eq:output}.
\end{enumerate}

\begin{equation}
\label{eq:forget}
F_t  = \sigma(W_F \dot [h_{t-1}, x_t] + b_F)
\end{equation}

\begin{equation}
\label{eq:input}
\begin{matrix}
I_t = \sigma(W_I \dot [h_{t-1}, x_t] + b_I)\\
\tilde{C_t} = tanh(W_C \dot [h_{t-1}, x_t] + b_C)
\end{matrix}
\end{equation}

\begin{equation}
\label{eq:cell}
C_t = F_t * C_{t-1} + I_t * \tilde{C_t}\\
\end{equation}

\begin{equation}
\label{eq:output}
\begin{matrix}
O_t = \sigma(W_O \dot [h_{t-1}, x_t] + b_O)\\
h_t = O_t * tanh(C_t)
\end{matrix}
\end{equation}

LSTM hidden states (as seen in figure~\ref{fig:lstm_cell}) can be connected to a softmax layer for text classification problems or it can be connected to another LSTM cell composing several LSTM layers. These LSTM layers would output a sequence of hidden state vectors, a vector for each input in a timestamp. This hierarchy of LSTMs layers enables a more complex representation of sequential data and it is often referred to as stacked LSTMs. In the next section, we discuss an architecture that often utilizes stacked LSTMs to generate language.
\subsection{Sequence-to-Sequence}
\label{sec:seq2seq}


\begin{figure*}[ht!]
  \includegraphics[width=\linewidth]{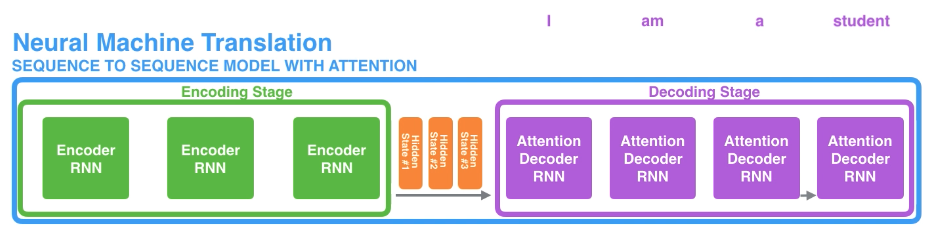}
  \caption{A Sequence to Sequence architecture is composed of an encoder and decoder LSTM networks. Source: \href{https://jalammar.github.io/visualizing-neural-machine-translation-mechanics-of-seq2seq-models-with-attention/}{Jalammar's Blog}.}
  \label{fig:seq2seq}
\end{figure*}

In this subsection, we describe the neural sequence-to-sequence (Seq2Seq) architecture. It is used to generate a sequence of tokens in a target language given a sequence of tokens from a source language. Seq2Seq learning maximizes the likelihood (or potentially other evaluation metrics~\cite{shen_minimum_2016}) of the target sequence (conversational response) given an input source sequence (user utterances).

Typically, Encoder-Decoder architectures using RNNs and applied to solve Seq2Seq tasks were first proposed by~\cite{cho2014learning} and~\cite{sutskever_sequence_2014}. In practice, gated RNNs such as LSTMs~\cite{hochreiter_long_1997} tend to perform better over vanilla RNNs~\cite{gu_incorporating_2016}.

Notably, Bahadanau et al.~\citeyearpar{bahdanau_neural_2014} propose an encoder-decoder architecture with attention, using a bi-directional LSTM as the encoder to transform an embedded source sequence $E = |e_1,...,e_{T_S}|$ into a vector $c$ of hidden states with equal length. This vector is known as the context vector or the thought vector. 

Each hidden state $h_t$ corresponds to an embedded token $e_t$ Each hidden state is computed by concatenating the hidden states of the forward and backward orders as follows:
\begin{equation}
\label{eq:encoding}
\begin{matrix}
\textbf{h}_t = [f(e_t,\textbf{h}_{t-1}) ; f(e_t,\textbf{h}_{t+1})] ;  \\ \\
\textbf{c} = \phi(\{\textbf{h}_1, ..., \textbf{h}_{T_S}\}) 
\end{matrix}
\end{equation}
where $\{h_t\}$ are hidden LSTM states, $f$ is a standard LSTM update function, and $\phi$ summarizes the hidden states. The decoder LSTM (recurrent neural network (RNN) architecture)~\cite{luong_effective_2015} generates the target token $a$ using the conditional probability defined in Eq.~\ref{eq:decoding} which takes in the context vector $c$ and generates target token $a$ at time $t$ using the following conditional probability:
\begin{equation}
\label{eq:decoding}
\begin{matrix}
\textbf{s}_t = f(a_{t-1},\textbf{s}_{t-1}, \textbf{c}) \\ \\
\: \: \: \: \: \:  p(a_t|a_{<t},E) = g( a_{t-1},\textbf{s}_t,\textbf{c}))
\end{matrix}
\end{equation}
where $s_t$ is the decoder LSTM state at time $t$, $a_t$ is the conversational response token at $t$ using function $g$.  $a_{<t}$ denotes previous predicted tokens $\{a_1,...,a_{t-1}\}$.\\

\begin{figure*}[ht!]
\centering
  \includegraphics[width=0.7\linewidth]{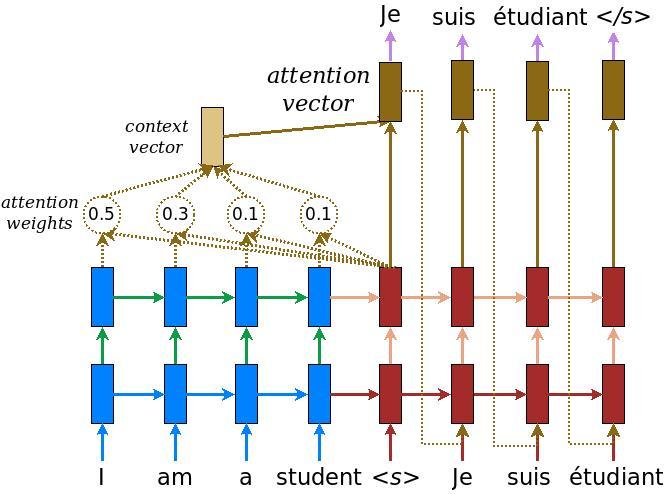}
  \caption{The sequence to sequence attention mechanism computing attention weights for each of the hidden states from the encoder network forming the context vector $c_t$  at time $t$ that is used by the decoder LSTMs. Figure adapted from~\cite{luong2015effective}. Adapted by: \href{https://medium.com/syncedreview/a-brief-overview-of-attention-mechanism-13c578ba9129}{SyncedReview on Medium}.}
  \label{fig:seq2seq_attention}
\end{figure*}
The context vector by itself is of a fixed length and hence it struggles with long sequences or sentences. To address this, the Seq2Seq model often comes with an attention mechanism. A noteable attention mechanism is the Bahdanau-style attention mechanism described in~\cite{bahdanau_neural_2014}, in which it considers a dynamic context $\textbf{c}_t$ in the decoding process.  This attender attends to the entire input space (i.e. soft attention) by representing $\textbf{c}_t$ as the weighted sum of the source hidden states as follows: 
\begin{equation}
\label{eq:attention}
\begin{matrix}
\mathbf{c}_t = \sum_{i=1}^{{T_S}} \alpha_{t,i}\: \textbf{h}_i \\ \\
\alpha_{t,i} = softmax(score(\textbf{s}_{t-1},\textbf{h}_i)) \\ \\
score(\textbf{s}_{t},\textbf{h}_i) = v_{z}^{\top} tanh(\textbf{W}_z[\textbf{s}_t; \textbf{h}_i])
\end{matrix}
\end{equation}

where $\alpha_{t,i}$ represents how well a target token $a_t$ and source token $e_i$ align, $T_S$ represents the last hidden state in $c$,  $s_t$ is the decoder LSTM state at time $t$, and lastly, $v_z$ and $W_z$ are weight matrices to be learned by the alignment model. The score $\alpha_{t, i}$ is parametrized by a feed-forward multi-layer perceptron neural network. Since the encoder uses a bi-directional LSTM each hidden state $\textbf{h}_i$ is aware of the context on both ends.  

In this section, we looked at the Seq2Seq model that uses an encoder LSTM network that processes input into a context vector and a decoder LSTM network that produces the output one token at a time. We noted how attention mechanisms weigh tokens based on how important they are at the current time step.
In the next section, we discuss a novel architecture, that still manages to perform well on very long sequences and is not as vulnerable to over-fitting as LSTMs.


\subsection{Transformers \& Transfer Learning}
\label{sec:transformers}
In this section, we discuss the Transformer architecture proposed by Vaswani et al.~\citeyearpar{vaswani_attention_2017}. In the previous section, we've touched on the attention mechanisms and how they supplemented RNNs in Seq2Seq models to model longer sequences. The Transformer architecture takes attention a step further to improve state of the art. In the infamous paper \textit{Attention is All you Need}, Vaswani et al.~\citeyearpar{vaswani_attention_2017} propose a new encoder-decoder architecture with a novel attention mechanism. The novel attention mechanism called the multi-head attention not only enables models to model longer sequences of text more effectively. The Transformer model also opens the door for transfer learning in natural language processing and generation. 
\newline
\newline
\begin{figure*}[ht!]
\centering
  \includegraphics[width=0.45\linewidth,keepaspectratio]{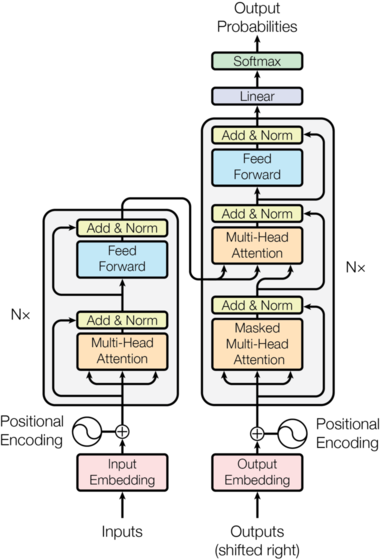}
  \caption{Overview of the Transformer architecture proposed in ~\cite{vaswani_attention_2017}. It is composed of embeddings, positional encoders, stacked encoders (left), and stacked decoders (right), }
  \label{fig:transformer_arch}
\end{figure*}

\textbf{Transfer learning} is a situation where what has been learned in one setting is exploited to improve generalization in another setting. Pan et al.~\citeyearpar{pan_survey_2010} define transfer learning as follows, consider a domain $D$ consisting of a feature space $\chi$  and a marginal probability $P(X)$ and $X = \{x_1,...,x_n\}$. Here, $x_i$ is a specific vector and $X\in \chi$ . Consider task $T$, consisting of a label space $\gamma$ and an objective function $P(\gamma|X)$. Transfer learning is the process aimed at improving target task $T_T$ in the target domain $D_T$ using knowledge from the source task $T_S$ in the domain $D_S$. Inductive learning is inferring mapping from a set of training data samples. The trained model then works with inductive bias, a set of assumptions related to the training data used to generalize on unseen data. Deep learning algorithms use inductive transfer techniques that utilize inductive biases of the source task to assist in the target task. To apply transfer learning, several methods are used: 1) Feature extraction, using pre-trained networks without its final layer as a feature extractor for other tasks. 2) Fine-tuning, where selected layers are re-trained from a pre-trained model, while other layers are frozen (weights are fixed). 3) Pretrained models, are models that are trained and perform well on a source task $T_S$, and the model weights and parameters are saved and then re-trained to perform a target task $T_T$. There are several types of transfer learning: 1) Domain adaptation, noted when there is a data shift between the source and target domains, usually the marginal probabilities $P(X_S) \neq P (X_T)$. An example of domain adaptation can be training a pre-trained sentiment classifier for movie reviews to classify product review sentiments. 2) Domain confusion aims to make the domain representations as similar as possible by applying pre-processing steps on the representations. The basic idea is to add another objective to the source model to encourage similarity between the domains by confusing the domain itself~\cite{ganin2016domain}. 3) Multitask learning, where the model receives information about multiple tasks simultaneously. 4) One and few-shot learning, where the model infers the required output based on one or a few training examples respectively~\cite{fei2006one}. 5) Zero-shot learning, where the model relies on no examples to learn a task.
\newline
\newline
\textbf{The Transformer architecture} is composed of 6 stacked encoders and 6 stacked decoders along with embeddings for both the inputs and outputs and positional encoders. Figure~\ref{fig:transformer_arch} overviews the proposed architecture.


We will first start by describing the encoders and decoders. The encoder receives inputs and passes them through the multi-headed attention. Multi-headed attention in each encoder is a type of self-attention. Self-attention helps the encoder map relationships between the current input token and other words in the input sequence as it encodes a specific word or a token~\cite{cheng2016long}. Then the output of the attention is fed into a feedforward (MLP) neural network. The output of each encoder is fed into another encoder until it reaches the last encoder, the sixth encoder. Each decoder receives the output of the sixth encoder and the target language input. It has both a feed-forward network and multi-headed attention but it also has another attention mechanism, the masked multi-headed attention. The masked multi-headed attention mechanism is also a self-attention mechanism, which takes in the output sequence as its input. It is modified to prevent attending to subsequent tokens/words so that predictions for the current timestep are only dependent on previously known output tokens. All the produced matrices are added with either the previous layer or the embedding and normalized according to~\cite{ba2016layer} in between as indicated in Figure~\ref{fig:transformer_arch}.

The multi-headed attention has an overall algorithmic complexity of $O(N^2)$ and is made up of 8 heads $h=8$ running in parallel. Each token in a sequence has an embedding vector. An embedding is a numerical form of a word that preserves its meaning by using dimensionality reduction and maintaining context. The paper uses 512 dimensions for the embedding size ($d_{model} = 512$). Each attention head takes in 3 input vectors for each input token: the query vector $Q$, the key vector $K$, and the value vector $V$. Each of these vectors has a dimension of 64. These vectors are computed by multiplying the embedded input token with 3 different weight matrices: $W^Q$, $W^K$, and $W^V$ respectively. The output of the attention head is computed as a weighted sum of the values vector. $d_k$ is the dimension of the key vectors which is 64.
\begin{equation}
\label{eq:transformers_attention}
Attention (Q,K,V) = softmax(\frac{QK^T}{\sqrt{d_k}})V
\end{equation}
To compute the attention for each head we start by computing the dot product of the query and the transposition of the key vector. The output is then scaled by $\sqrt{d_k}$ which is 8. Next, the softmax function is applied so that all the values are between 0 and 1. Lastly, we compute the dot product of the softmax output with the values vector. The process is called the Scaled Dot-Product Attention and is described in Equation~\ref{eq:transformers_attention}. 

To compute the multi-head attention we take the output from each attention head, concatenate them, and multiply them with a bigger weight matrix $W^O$ as described in Equation~\ref{eq:transformers_multiattention}.

\begin{equation}
\label{eq:transformers_multiattention}
\resizebox{0.9\linewidth}{!}{$MultiHead (Q,K,V) = Concat(head_1,...,head_h)W^O$}
\end{equation}

The weight matrices  $W^Q$, $W^K$, $W^V$, and $W^O$ are all learned and updated throughout the training process which greatly assists in the process of transfer learning. Each of the 8 attention heads can focus on a different relationship between the input words (e.g., one for pronoun-noun, one for gender, etc.)

Another component is positional encoding applied to both the input and output sequences. The positional encoding has a dimension of 512 which is the same as the embeddings. Positional encodings provide a vector to be added to the input embedding. This provides fixed positioning of tokens during training which allows further extrapolation of longer sequences. The paper uses sinusoidal version of positional embeddings described in Equation~\ref{eq:transformers_pe}, where $pos$ is the position and $i$ is the dimension. Each dimension of a positional encoding corresponds to a sinusoid which goes from $2\pi$ to 10000.

\begin{equation}
\label{eq:transformers_pe}
\begin{matrix}
PE_{(pos,2i)} = sin(pos/10000^{2i/512})\\
PE_{(pos,2i+1)} = cos(pos/10000^{2i/512})
\end{matrix}
\end{equation}
There are many applications of the Transformer model and it is not restricted to the field of NLP. We note in particular BERT~\cite{devlin2018bert} which advanced state of the art for language modeling and GPT-3~\cite{brown2020language} which advanced language generation. In the next section, we will discuss the GPT models.


\begin{figure*}[ht!]
\centering
  \includegraphics[height=7.5cm,keepaspectratio]{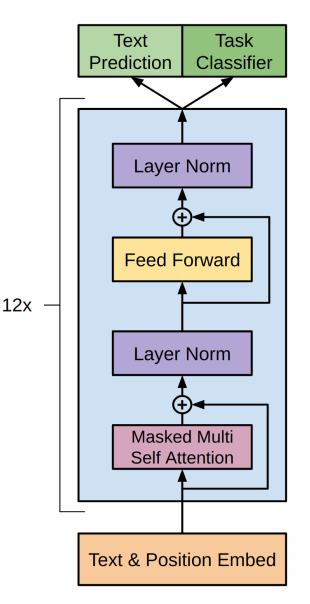}
  \caption{The GPT-1 architecture proposed in ~\cite{radfordimproving}. It is composed of 12 stacked decoders. Each decoder containing self-attention, followed by a position-wise feed-forward network with normalization in between. }
  \label{fig:gpt_1}
\end{figure*}

\subsection{GPT Models}
In the section, we describe a specific transformer architecture, the Generative Pretrained Transformer family of models.
\label{sec:gpt}

\subsubsection{GPT-1}
Prior to GPT-1 most NLP models required annotated corpora for natural language understanding. Annotated data for specific natural language understanding (NLU) tasks are scarce while large unlabeled corpora are abundant. GPT-1, or Generative Pre-trained Transformer 1, introduces the concept of generative pre-training on large unlabeled text and then fine-tunes for specific NLU tasks~\cite{radfordimproving}. GPT-1 demonstrates significant gains in several NLU tasks and marks a significant step towards the trend of task-agnostic NLP models. Next, we describe the semi-supervised learning process (pre-training then fine-tuning) more in detail.

The unsupervised pre-training is a language model is tasked with maximizing the following objective for a corpus of tokens $U = {u_1,...,u_n}$: 
\begin{equation}
\label{eq:gpt_pretraining}
L_1(U) = \sum_i log P(u_i|u_{i-k},...,u_{i-1};\Theta)
\end{equation}

where $k$ is the size of the context window, P is a conditional probability modeled by a neural network with parameters $\Theta$. The parameters are trained using stochastic gradient descent (SGD) described earlier in~\ref{sec:mlp}.

The fine-tuning training (supervised training) takes in a labeled dataset $C$ with a sequence of tokens ${x_1,...,x_m}$, and a label $y$ for the token sequence and aims to maximize the objective given the pre-training objective $L_1$, and a weight $\lambda$ set to 0.5 in the paper:
\begin{equation}
\begin{matrix}
\label{eq:gpt_finetuning}
L_2(C) = \sum_{(x,y)} log P(y|x_1,...,x_m)\\
L_3(C) = L_2(C) + \lambda L_1(C)
\end{matrix}
\end{equation}
Using the language modeling (pre-training) objective helps with improving the supervised model's generalization and accelerating convergence~\cite{radfordimproving}.
\newline
\textbf{The GPT Architecture } is composed of 12 stacked transformer decoder layers (see the decoder component in Figure~\ref{fig:transformer_arch}), with multi-headed masked self-attention. The decoder layer had 12 attention heads, the embedding size is 768 ($d_{model} = 768$), an Adam optimizer~\cite{kingma2014adam} was used with a learning rate of $\alpha = 2.5 \times 10^{-4}$, and a drop out rate of 0.1. As shown in Figure~\ref{fig:gpt_1}, the model applies its multi-headed masked self-attention on the input tokens. Following attention, are position-wise feed-forward layers generate a distribution over the target tokens. The dimension of the position-wise feed-forward layer is 3072. During pre-training, the output distribution $P(u)$ for a token $u$ is computed through the following:
\begin{equation}
    \begin{matrix}
    h_0 = UW_e + W_p \\
    h_i = \textrm{decoder}(h_{i-1})\forall{i} \in{[1,n]} \\
    P(u) = \textrm{softmax}(h_nW^T_e)
    \end{matrix}
\end{equation}
Here, $U = {u_{k},...,u_{1}}$ is the context vector of tokens starting from $k$ the size of the context window, $n$ is the number of layers (12 layers), and $W_e$ is the token embedding matrix, and $W_p$ is the position embedding matrix. We observe that the first decoder layer $h_0$ takes in the embedding matrix and position embedding matrix. Subsequent decoder layers take in the previous layer, and softmax is applied at the last transformer layer and maximizes for Equation~\ref{eq:gpt_pretraining}. When fine-tuning GPT-1, input tokens $x^1,...,x^m$ which come along with label $y$ are passed through the pre-trained model to obtain the last decoder activation $h^m_l$. It is then passed to the linear layer (shown in Figure~\ref{fig:transformer_arch}) with parameters $W_y$ to predict $y$ and maximizes for Equation~\ref{eq:gpt_finetuning} :
\begin{equation}
    P(y|x^1,...,x^m) = \textrm{softmax}(h^m_l W_y)
\end{equation}
The model was trained for 100 epochs with mini-batches of 64 and a sequence length (context window) of 512 and used GELU (Gaussian Error Linear Unit)~\cite{gelu} as its activation function. Byte Pair Encoding (BPE)~\cite{sennrich-etal-2016-neural} with a vocabulary size of 40,000 was used. GPT-1 had 117 million parameters in total. The Bookcorpus dataset~\cite{zhu2015aligning} was used for the pre-training stage. GELU is approximated by the following using $tanh$ described in section~\ref{sec:mlp}:
\begin{equation}
\begin{split}
    GELU(x) = 0.5x(1 + tanh[ \\ \sqrt{2/\pi}(x + 0.044715x^3)])
    \end{split}
\end{equation}

\begin{figure*}[!htbp]
\centering
  \includegraphics[width=\linewidth]{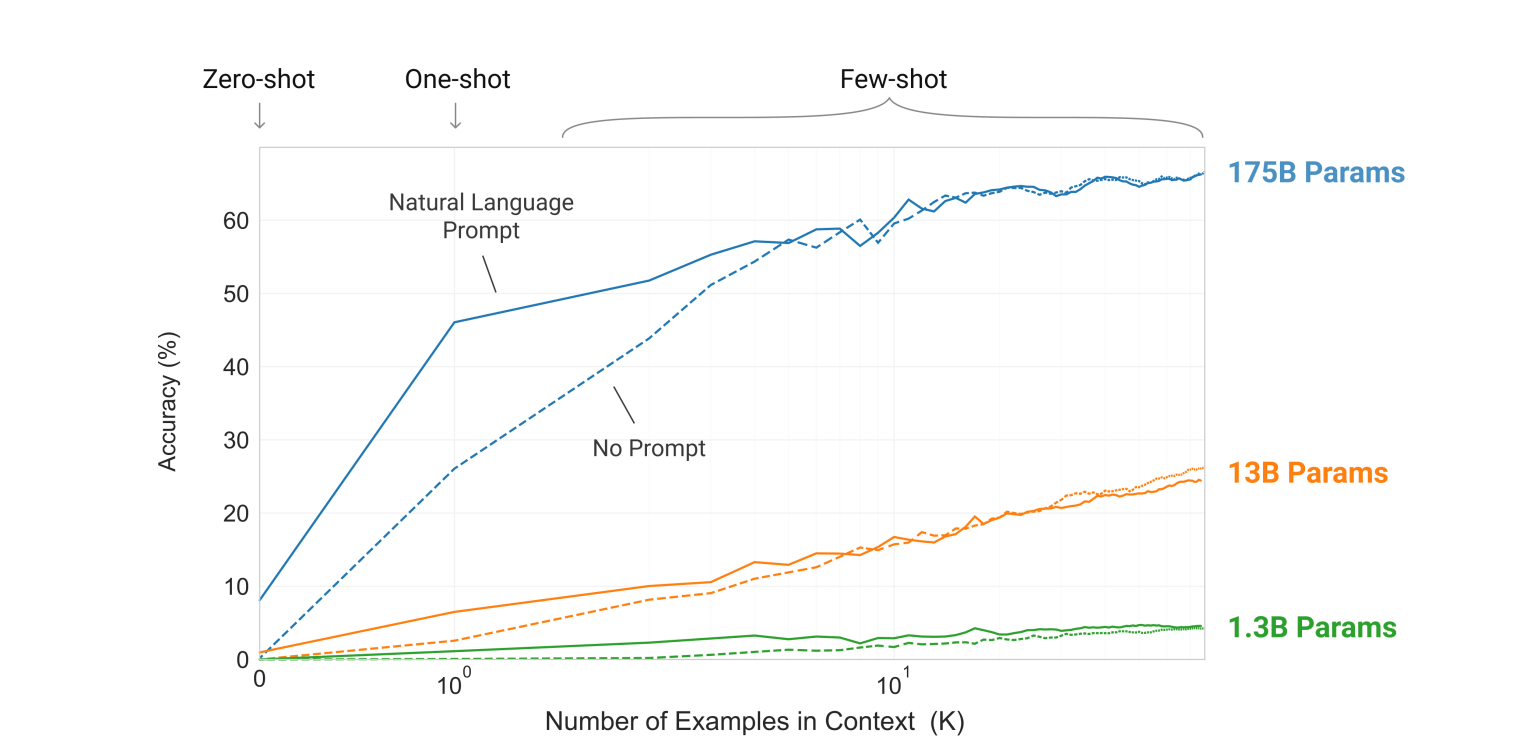}
  \caption{Figure from ~\cite{brown2020language}. Shows GPT-3 with different number of parameters making more use of context (K) on a simple task requiring the model to remove random symbols from a word. }
  \label{fig:gpt3_params}
\end{figure*}

\subsubsection{GPT-2}
The Generative Pre-trained Transformer 2 (GPT-2) serves as the first attempt at creating a pre-trained language model that can perform downstream tasks without any fine-tuning or architectural change. This is also known as the zero-shot setting. To achieve this goal, GPT-2 updates the objective function  from the pre-training objective $P(output|input)$ to include a particular task $P(output|input,task)$~\cite{radfordlanguage}. This is called task conditioning. With task conditioning GPT-2 can produce different outputs for the same input given a different task. The task is passed as a natural language description (sequence of tokens describing the task). GPT-2 is trained using a dataset called WebText\footnote{\url{https://github.com/jcpeterson/openwebtext}} which is around 40 GB in size~\cite{radfordlanguage}. GPT-2 experiments highlighted that the perplexity (intrinsic evaluation) of language models on the same dataset decreases the bigger the language model in terms of parameters. 
\newline
\textbf{The GPT-2  Architecture } follows the architecture of GPT~\cite{radfordimproving} with a few notable differences. GPT-2 is made up of 48 stacked transformer decoder layers. The normalization layers were moved to the input of the self-attention and feed-forward sub-blocks. There is also an additional normalization layer was added after the final self-attention block. Weight initialization was also modified by a factor of $1/\sqrt{N}$ where $N$ is the number of residual layers. The word embedding dimension was increased to 1600 ($d_{model} = 1600$). Byte Pair Encoding (BPE)~\cite{sennrich-etal-2016-neural} with an increased vocabulary size of 50,257 was used. The model was trained with a larger batch size of 512 and a larger context window of 1024. This resulted in a total of 1.5B parameters for GPT-2 which is almost 10 times bigger than GPT-1.

\subsubsection{GPT-3}
Large transformers such as BERT~\cite{devlin2018bert}, and GPT-2~\cite{radfordlanguage} are task agnostic in architecture, however, they require fine-tuning on specific tasks using thousands of examples. In a new transformer-based architecture, researchers at OpenAI explore greatly scaling up transformer-based language models to explore its task-agnostic performance using few-shot and zero-shot learning.  GPT-3, or Generative Pre-trained Transformer 3, is a massive auto regressive transformer-based model that underwent unsupervised pre-training on large corpora (over 300 billion tokens of text or 45 TB). The data sources include 3 years of filtered CommonCrawl~\cite{wenzek2020ccnet}, Wikipedia, and WebText. GPT-3's pre-training objective is to predict the next word given a sequence of words. GPT-3 is massive, containing about 175 billion parameters, making it one of the world's largest pre-trained transformer models for language generation in the world. GPT-3 is used without gradient updates nor fine-tuning but rather with few-short demonstrative interactions with the model. Researchers have tested the model on various NLP tasks including question answering, translation, and cloze tasks competing with state-of-the-art models without fine-tuning. GPT-3 generates language one token at a time, just like previously mentioned models, given a sequence of tokens as input in many different domains. Upon its release to the public enthusiasts have used GPT-3 for many different applications. These applications include mainstream tasks such as translating into different languages, writing google ads and has been particularly used to generate code from natural language such as SQL, Bash, and Python.  For a thorough list of GPT-3 applications please refer to this\footnote{\url{https://github.com/elyase/awesome-gpt3}}. 
\newline
\textbf{The GPT-3 Architecture} is the same as that of a GPT-2 model~\cite{radfordlanguage}. Similar to the GPT-2~\cite{radfordlanguage}, the  GPT-3 model is composed of stacked transfomer Decoder layers (see the decoder component in Figure~\ref{fig:transformer_arch}) and it includes the modified initialization, pre-normalization, and reversible tokenization. GPT-3 is different from GPT-2, in that it uses alternating dense and banded sparse attention patterns in the transformer layers proposed in the Sparse Transformer paper~\cite{child2019generating}. Open AI researchers have experimented with different GPT-3 configurations, sizes, and parameters. An example experiment is shown in Figure~\ref{fig:gpt3_params}. The final GPT-3 model uses 96 transformer decoder layers, 96 attention heads, the dimension of the model ($d_{model} = 12888$), dimension of each attention head ($d_{head}= 128$), batch size is 3.2M, and $\alpha = 0.6 \times 10^{-4}$.

\section{Code Intelligence (CI)}

\label{sec:plp}
With the availability of \textit{big code} from public sources like GitHub and StackOverflow, an opportunity to use data-driven tools and algorithms comes to light. This opens the door for machine learning, and NLP techniques, in particular, to be used for various applications on source code. Such applications tend to improve the software development process and are gaining attention from both  AI and software engineering academics. Applications of machine learning and NLP techniques on source code includes applying statistical models~\cite{oda_learning_2015}, neural networks such as LSTMs~\cite{rabinovich_abstract_2017}, and more recently pre-trained transformers including CodeBERT~\cite{feng_codebert_2020}, CodeGPT~\cite{svyatkovskiy2020intellicode}, Codex~\cite{chen2021evaluating} which are respectively variants of BERT~\cite{devlin2018bert}, GPT-2~\cite{radfordlanguage}, and GPT-3~\cite{brown2020language} pretrained on source code . Recently, researchers at Microsoft have published an international benchmark CodeXGLUE~\cite{CodeXGLUE} for Code Intelligence, similar to existing NLP benchmarks such as GLUE~\cite{wang2018glue} and SQuaD~\cite{rajpurkar2018know, rajpurkar2016squad}. Another benchmark worth mentioning is GLUECode~\cite{karmakar2020gluecode}. GLUECode evaluates machine learning models on local and global understanding evaluation of source code. Different from CodeXGLUE, GLUECode contains two tasks for ``local reasoning" in which models are tasked with showing understanding of codes on a local level. The first task is operator prediction, in which a model is tasked with predicting a masked operator including boolean comparison operators and arithmetic operators. The second task is NPath prediction, where models are tasked with counting the number of distinct paths control flow can take in a method. We categorize works in NLP on source code fall into a set of areas. The areas include Program Synthesis, Program Analysis, Program Repair, Clone  Detection, Defect Detection, Cloze Testing, Code Translation, Code Refinement, Code Completion, Code Search, Code Documentation Generation, Documentation Translation, Semantic Parsing, and Code Generation. This section will overview each of these areas and highlight notable works. We focus on Code Generation and semantic parsing more in particular for this survey.
\subsection{Program Synthesis}
Artificial intelligence research has long aimed at having computers synthesize their programs~\cite{manna1971toward,waldinger1969prow}. Program Synthesis involves constructing full or partial programs that complete a specified task~\cite{allamanis_survey_2018,gulwani2017program}. Program specifications are taken by the program synthesizer model as input and an executable program is synthesized by the model as the output. Program specifications are traditionally expressed with formal logic statements. Other program specifications include providing the model with example input/output pairs and more recently natural language descriptions of what the program needs to accomplish. In program synthesis, if the program specifications include natural language descriptions it is accompanied by another specification type such as input/output pairs ~\cite{austin2021program,shi2020tf,ye2020optimal}. If the program specification is in natural language only then it would be considered a semantic parsing task which is described in \S{\ref{subsection:semantic_parsing}}. Program synthesis works usually synthesis programming using restricted domain-specific languages (DSLs)~\cite{gulwani2011automating} or languages that have more features but are still designed with program synthesis in mind~\cite{odena2020learning,ellis2021dreamcoder}. More recently, program synthesizers have started to synthesize programs in general-purpose languages such as Python ~\cite{shi2020tf,austin2021program}.

Program synthesis approaches include applying both symbolic ~\cite{manna1975knowledge, manna1980deductive} and neuro-symbolic ~\cite{parisotto2016neuro} techniques~\cite{balog2016deepcoder,odena2020learning,devlin2017robustfill}. Symbolic approaches are also known as rule-based program synthesis~\cite{manna1980deductive} focus on deriving a program from a well-defined specification using a set of formal grammar rules. A drawback of the symbolic approach is it often relies on using formal logic to describe the program specification, however writing the formal logic to describe the program is often more difficult than writing the program itself. To remedy this issue, modern program synthesis models often learn from input/output examples. A recent notable work utilizing input/output examples in various domains is \textit{DreamCoder}~\cite{ellis2021dreamcoder}. DreamCoder learns to synthesize programs in a domain-specific language using input and output examples to solve inductive programming tasks such as list processing and text editing, creative tasks such as drawing pictures and building scenes and to solve classical physics problems. DreamCoder uses a wake-sleep learning algorithm. In its, sleep DreamCoder goes through to learn concepts from combinations of DSL primitive operations improving its library of operations in the process. When DreamCoder is awake it uses operations from its library to solve problems given input and output examples. There are also notable works that synthesize programs in Python, a general-purpose programming language, which has the potential to power tools to further enable programmers. Austin et al.~\citeyearpar{austin2021program} explore the limits of using a large pretrained transformer-based language model to synthesize basic python programs to solve entry-level programmer tasks and math problems given input/output examples and a natural language description. The authors note that large language models enable program synthesis models to consider various types of program specifications such as natural language. Furthermore, the authors demonstrate that such language models can reduce the error rate of synthesized programs by up to 50\% when engaging in dialogue with a human and incorporating human feedback given on the synthesized code. Shi et al.~\citeyearpar{shi2020tf} propose \textit{TF-Coder}, a program synthesizer for TensorFlow, a well-known deep learning library by Google. \textit{TF-Coder} uses input/output examples and natural language description as program specifications and generates Python code in the TensorFlow library to solve real-world tensor manipulation problems achieving superhuman performance.

Program synthesis models are generally evaluated based on how many compilable (executable) programs are synthesized. They are also evaluated based on how many programs are logically correct, i.e. match the specification constraints or solve the programming problem by passing pre-programmed unit tests~\cite{austin2021program}. Other metrics also include how long it takes for the model to synthesize the program, and duration speedup compared to human programmers. Recent work by Schuster et al.~\citeyearpar{schuster2021programming} proposes a dataset, programming puzzles (P3), to evaluate program synthesis models. Program synthesis models are tasked with predicting an input $x$ to a short Python program $f$ with makes $f$ output True.

Although program synthesis generally assumes that the program compiles with some specification, some notable works utilize program synthesizers to generate random but functioning programs for benchmarks and compiler fuzzing. Cummins et al.~\citeyearpar{cummins2017synthesizing} synthesize a large number of programs used as benchmarks in OpenCL code. Fuzzing is a popular method that creates structured data samples that are used as inputs to test complex software such as compilers and web applications. Patra and Pradel~\citeyearpar{patra2016learning} synthesize programs using \textit{TreeFuzz} in JavaScript using a corpus of example structured data (in this case example program outputs). These programs generate random structured data for fuzz-testing JavaScript interpreters. Program Synthesis has been applied to build usable products in fields including data science~\cite{drosos_wrex_2020}, to assist data scientists in wrangling data using Python, and general software engineering to generate code edit suggestions based on repetitive changes~\cite{miltner_fly_2019}.


\subsection{Program Analysis}
\label{subsec:program_analysis}
Computers can execute programs but they do not necessarily understand what the programs do or mean. When analyzing program source code, we can computationally estimate a program's behavior, functionality, complexity, meaning, etc. Program analysis focuses on extracting semantic properties from programs. In a sense, it is similar to natural language understanding (NLU) in natural language processing where both fields focus on comprehending a snippet of. Probabilistic models of source code compose a notable family of models to analyze programs and extracting properties. Sometimes codebases contain uninformative variable names such as \texttt{e} and which makes it difficult for humans to comprehend code. This can be remedied by refactoring codebases to include descriptive comments and variable names. Raychev et al.~\citeyearpar{raychev2015predicting} propose a system \texttt{JSNice} that models the statistical patterns of source code through an AST-based dependency network to predict variable names and types in JavaScript programs. A notable issue in program analysis has been scalable but imprecise models. Models that tended to scale on large datasets tended to have higher numbers of false positives. To remedy this, Oh et al.~\citeyearpar{oh2015learning} proposes using Bayesian Optimization to optimize models for static program analysis and report achieving high precision while lowering false positives. Mangal et al.~\citeyearpar{mangal2015user} take a more user-centered approach and utilize online learning to tune their model. The user would provide feedback to the program analysis system as to whether the user likes or dislike the program analysis output. Mapping comments onto code tokens can be a useful feature in helping programmers debug and understand code. In an attempt to model this, Panthaplackel et al.~\citeyearpar{panthaplackel2020associating} work on associating entities mentioned in Javadoc comments with source code tokens using a feedforward neural network trained as a classifier trained on noisy data from Github. Managing large software repositories can be gruesome. Source code classification and tagging are crucial in helping to organize, search, and reuse codebases. In an attempt to automate source code tagging and classification, Mou et al.~\citeyearpar{mou_convolutional_2016} propose a novel architecture of Tree-based convolutional neural networks (TBCNNs). TBCNNs combine structured information from Abstract-Syntax Trees (ASTs), which are tree-based meaning representations of code snippets, with Convolutional Neural Networks (CNNs). CNN's are a deep learning model derived from MLPs. Their work is then applied to classify code bases by functionality with over 104 unique functionalities. TBCNNs were reported to outperform other classification machine learning models including SVM and RNNs for this task. Program analysis has been also useful in software reverse engineering, in particular, recovering variable names when decompiling C code~\cite{lacomis2019dire}. 


\subsection{Clone Detection}
Code Clone Detection is another area of research that focuses on detecting portions of code that mean or do the same thing. Program analysis (see~\S{\ref{subsec:program_analysis}}) focuses on analyzing semantic properties of programs. Clone detection is closely related to program analysis and can be considered a sub-field of program analysis that is particularly focused on estimating semantic similarity between two codes. Early work on code clone detection by Chilowicz and Duris~\citeyearpar{chilowicz2009syntax} focused on indexing and retrieving abstract syntax trees. Clone detection has been approached as a machine learning problem of binary classification of two codebases (e.g. labeling a code pair with 1 to indicate that the pair is a clone and labeling with 0 to indicate otherwise)~\cite{svajlenko2014towards,wei2017supervised,wang2020detecting,wang2021mulcode}. Code clone detection has also been approached as an information retrieval problem, where the model is tasked with retrieving semantically similar codes to a given code input~\cite{ganguly2018retrieving, mou2016convolutional,CodeXGLUE}. Code clone detection is included as a task on the CodeXGLUE benchmark~\cite{CodeXGLUE}. The task evaluates models using the F1-score for the binary classification subtask and Mean Average Precision (MAP)~\cite{beitzel2009,CodeXGLUE} for the code retrieval task. Mean Average Precision (MAP) is the arithmetic mean of the average precision values ($AP$) for a system over a set of $n$ documents. MAP is described in equation~\ref{eq:map}. 

\begin{equation}
\label{eq:map}
    MAP = \frac{1}{n}\sum_{n}AP_{n}
\end{equation}

Notable datasets for clone detection include the POJ-104 dataset~\cite{mou2016convolutional} for clone code retrieval. The POJ-104 dataset is collected from a programming open judge (OJ) system where students submit their source code for 104 problems. The dataset contains about 500 student submissions in the C/C++ programming language for each problem. The task is to retrieve other codes that solve the same problem as the input example. Another notable dataset is the BigCloneBench dataset~\cite{svajlenko2014towards} which focuses on the binary classification of clone codes. After filtering the dataset contains about $1,731,860$ Java code pairs. 

Currently, to the best of our knowledge, CodeBERT~\cite{feng_codebert_2020} outperforms prior models on both the POJ-104 and the BigCloneBench datasets. Prior works in clone detection have used AST-based RNNs with pretrained word embeddings like word2vec~\cite{buch2019learning}. Other notable approaches for clone detection include Clone Detection with Learning to Hash (CDLH)~\cite{wei2017supervised} which learns hash functions and representations of code snippets via an AST-based LSTM. Another approach is Flow-augmented abstract syntax tree (FA-AST), which augments the original AST representation of source code with data flow edges and uses graph neural networks~\cite{wang2020detecting}. MulCode~\cite{wang2021mulcode} integrates a pre-trained transformer model BERT and an AST-based LSTM to encode a code sequence and structure and is evaluated on three classification tasks including clone detection.

Code clone detection has become of increasing importance since the rise of machine learning on ``big code". Lopes et al.~\citeyearpar{lopes2017dejavu} discovered that most of the code on Github is near-duplicate. A study followed by Allamanis~\citeyearpar{codeduplication2019} explores the effects of code duplication on machine learning models for source code. Allamanis observes big inflation in machine learning metrics when testing on duplicated code corpora. Code clone detection is crucial in creating higher quality code datasets for machine learning and can also have other uses such as detecting code plagiarism in academic contexts~\cite{ganguly2018retrieving}.

\subsection{Code Search}
\label{subsec:codesearch}


 \begin{table*}[!htbp]
 \fontsize{9}{11}\selectfont

\renewcommand{\arraystretch}{1.5}
\begin{tabular}{p{4cm} p{6.5cm} ccc }
\toprule


\multicolumn{1}{c}{\textbf{Model}}   & \multicolumn{1}{c}{\textbf{Description}} & \multicolumn{1}{c}{\textbf{Best NDCG}} & \multicolumn{1}{c}{\textbf{Baseline}} & \multicolumn{1}{c}{\textbf{GitHub}}   \\ \toprule 


Neural Bag of Words (BoW)   &  Encodes the codebases and English queries by encoding them into a standard token embedding with masking. The model optimizes to minimize the cosine distance between the code and the query.  & 0.384 & 0.340 & \cite{wuchen2020} \\

Self-Attention & Transformer masked multi-headed attention~\cite{vaswani_attention_2017} is used to compute  the token embeddings. &  0.240  & 0.240 & - \\

One Dimensional Convolutional Network (CNN) & CNN is used over query and code embeddings. & 0.165 & 0.165  & - \\

   Bidirectional RNN & Employing gated recurrent network (GRU) to summarize the query. & 0.046  & 0.046  & -\\ 
\bottomrule

\end{tabular}

\caption{Overview of top scoring CodeSearchNet models sorted by the best score for the particular model across all leaderboard submissions. The score is the average Normalized Discounted Cumulative Gain (NDCG) score across all six programming languages.}
\label{tab:codesearch}

\end{table*}

 Searching for code snippets online is a common activity for programmers and software engineers.  Semantic code search is the task of retrieving semantically equivalent code to a natural language query~\cite{husain_codesearchnet_2019}. Husain et al.~\citeyearpar{husain_codesearchnet_2019} release a corpus specific for the task of code search called CodeSearchNet\footnote{\url{https://github.com/github/CodeSearchNet}}. The corpus is collected from public, licensed, and non-forked GitHub repositories. Functions and methods are parsing using \textit{TreeSitter} and corresponding documentation is parsing using regular expressions. The training set of the CodeSearchNet corpus includes only code functions that have an associated documentation string with them. Other inclusion and exclusion criterion include: the documentation is shortened to only the first paragraph, samples that include 3 tokens or less of documentation or 3 lines or less in its function body are removed. Additionally, methods with the string \texttt{test} in its function name, constructors and standard extension methods such as \texttt{\_\_str\_\_} are removed, and lastly duplicate functions are removed from the dataset using clone detection methods described in~\cite{lopes2017dejavu,codeduplication2019}\footnote{\url{https://github.com/Microsoft/near-duplicate-code-detector}}. It contains around 6 million functions in six programming languages: Java, Go, JavaScript, Python, Ruby, and PHP along with corresponding 2 million natural language queries in English. While the dataset is relatively big, it suffers from being quite noisy. Scraped function documentation is far from a query, hence the authors reported difficulty in identifying how accurately a documentation sample describes its corresponding code function. The code search task is evaluated using Mean Reciprocal Rank (MRR) and Normalized Discounted Cumulative Gain (NCDG) a commonly used metric in information retrieval described in~\cite{vechtomova2009introduction}. Top scoring attempts of each unique method are listed in Table~\ref{tab:codesearch} along with the GitHub reference, and the baseline score for that approach. Submissions without publications or with private code bases were excluded. A notable attempt not mentioned in the table by Arumugam~\cite{arumugam2020semantic} who uses code2vec~\cite{alon2019code2vec}, a code embedding of abstract syntax tree paths for semantic code search. He uses code2vec to embed the code functions and combined it with another neural model such as Neural Bag of Words (NBoW) which are table-look up embeddings or embeddings with self-attention for the query and scores 0.148 average NDCG using NBoW for the query and code2vec for code.

 There are additional works that tackled the problem of code search that do not use the CodeSearchNet dataset exclusively. Reid et al.~\citeyearpar{reid2020optimising} propose NLP2TestableCode an Eclipse IDE plugin that retrieves code snippets from the StackOverflow database using natural language queries and integrates  the retrieved code into a Java codebase. Li et al.~\citeyearpar{li2019neural} propose a dataset composed of natural language queries and code snippets for evaluating code search models. The dataset was constructed using 287 Stackoverflow question and answer pairs as well as from method code snippets from the most popular Android repositories on Github that correctly answer the Stackoverflow question. The Github corpora contains about 24,549 repositories and 4,716,814 methods.
 
 \subsection{Code Documentation Generation \& Summarization}
 \label{subsec:comment_gen}
 The goal of code documentation generation is to generate a comment in natural language that accurately describes a given code snippet. This research area intersects heavily in terms of datasets and models with Code Generation (\S\ref{subsec:code_gen}) since both areas heavily utilize corpora of source code and natural language and utilize machine translation approaches. Notable work by Oda et al.~\citeyearpar{oda2015learning} propose two novel dataset of Python and natural language comments called the Django and Euler datasets. The authors then statistical machine translation to generate comments in natural language given a Python source code snippet. The system was evaluated using BLEU~\cite{papineni2002bleu} and human evaluation of the generated comments. Fudaba et al.~\citeyearpar{fudaba2015pseudogen} build a tool called \textit{Pseudogen} that also generates pseudocode in both Japanese and English from Python code using statistical machine translation.  Iyer et al.~\citeyearpar{iyer2016summarizing} propose a dataset and an LSTM-encoder decoder model to produce sentences in natural language that describe \texttt{C\#} code snippets. Gros et al.~\citeyearpar{gros2020code} examine the task of generating comments in 4 code-natural language datasets using a sample information-retrieval (IR) model using the BM25 scoring function~\cite{robertson1994some,amati2009}, LSTM-based sequence-to-sequence, and a transformer-based sequence-to-sequence model. The authors find that the simple IR model provides competitive performance on these datasets. Notable work by Gao et al.~\citeyearpar{gao2020stack} propose generating Stack Overflow questions given source code which is similar to generating comments. Code Documentation Generation is also included as a task in the CodeXGLUE~\cite{feng_codebert_2020} which utilizes the CodeSearchNet corpus~\cite{husain_codesearchnet_2019} for training and testing machine learning models. More recent approaches to code documentation utilize pretrained transformers. Currently, CoTexT~\cite{phan2021cotext} outperforms CodeBERT~\cite{feng_codebert_2020}, PLBART~\cite{ahmad-etal-2021-unified}, and ProphetNET-X~\cite{qi2021prophetnet} in this task. The CodeSearchNet corpus is further described in \S\ref{subsec:codesearch} and the models are described later on in the code generation subsection~\S\ref{subsec:code_gen}. Code summarization focuses on generating method names in natural language given a method's body of code snippets.A notable work on representing codes as vectors is code2vec~\cite{alon2019code2vec} which focuses on embedding abstract syntax tree paths into a fixed-length vector. Code2seq~\cite{alon2018code2seq} is an LSTM-based sequence to sequence model that encodes the raw source code tokens along with the AST paths in its encoder and attends to the relevant paths while decoding. Code2seq outperforms code2vec on code summarization and outperformed prior state of the art in code documentation generation CodeNN~\cite{iyer2016summarizing} by ~2\% BLEU.
 
\subsection{Documentation Translation}
Code Documentation Translation focuses on translating web-pages containing code-related documentation such as library docs from a source language into a target language. CodeXGLUE proposes this task and curates a dataset by crawling the Microsoft Documentation website\footnote{\href{https://docs.microsoft.com/en-us/}{https://docs.microsoft.com/en-us/}}. and it includes software and code description documents in various languages~\cite{CodeXGLUE}. The task focuses on translation between English and other languages such as Danish, Latvian, Norwegian, and Chinese. The multi-lingual zero-shot transformer model proposed by Johnson et al.~\citeyearpar{johnson2017google} and a variant of the model with the encoder initialized with XLM-R~\cite{conneau-etal-2020-unsupervised} pretrained weights were used as models for this task. The pretrained transformer was reported to outperform the baseline on all translation tasks~\cite{CodeXGLUE} using the token overlap metric BLEU~\cite{papineni2002bleu}. The pretrained transformer scored 66.16 while the transformer baseline score 52.67 BLEU.

\subsection{Semantic Parsing}
\begin{figure*}[h!]
\centering
 \begin{subfigure}[h]{\textwidth}
    \centering
    \includegraphics[height=5cm,keepaspectratio]{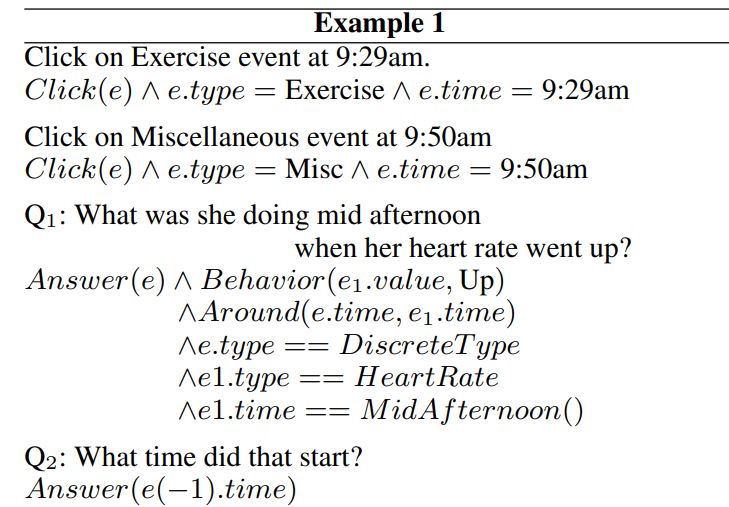}
    \caption{Sample task of parsing natural language interactions into a domain specific logical form~\cite{chen2019context}.}
    \label{fig:calculus}
  \end{subfigure}\par\medskip
\begin{subfigure}[h]{0.5\textwidth}
    \centering
    \includegraphics[height=1.25in]{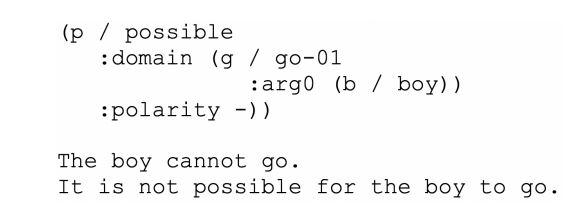}
    \caption{Abstract Meaning Representation of two sentences that are semantically equivalent but have different grammar structures~\cite{banarescu2013abstract}.}
    \label{fig:amr}
  \end{subfigure}%
~  
  \begin{subfigure}[h]{0.5\textwidth}
    \centering
    \includegraphics[height=1.3in]{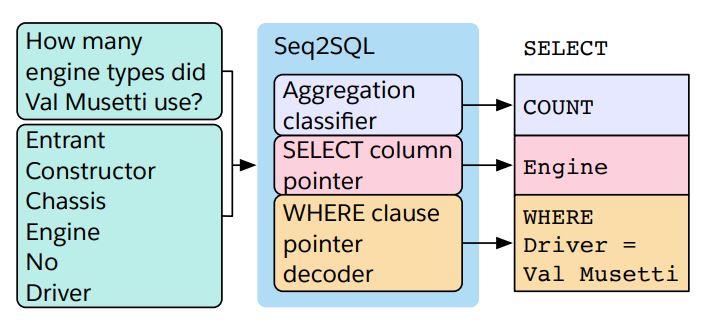}
    \caption{Sample task of parsing natural language into an SQL query~\cite{zhong_seq2sql_2017}.}
    \label{fig:sql}
  \end{subfigure}\par\medskip
  \caption{Example semantic parsing tasks with various meaning representations.}
  \label{fig:semantic_parsing}
  
\end{figure*}

\label{subsection:semantic_parsing}
When interfacing with machines via natural language, a key component is the machine's understanding of the natural language intent. The intersection between natural language processing, information retrieval, and interaction with humans is called natural language understanding (NLU). Semantic parsing converts natural language utterances to executable meaning representations, such as lambda calculus or database queries, that can be executed on a knowledge-base~\cite{kamath_survey_2019}. Semantic parsing can also be known as auto-formalization, which is defined as the task of turning informal descriptions to formal and automatically verifiable formats~\cite{szegedy2020promising}. Semantic parsing has become increasingly prominent in applications focused on interfacing via natural language with computer systems~\cite{yin_towards_2020}. Semantic Parsing enables machines to identify roles and objects in a sentence~\cite{fernandez2020transition}, converting natural language to database queries~\cite{zhong_seq2sql_2017} that can then be plugged into conversational assistants like Alexa, interfacing with robots using natural language commands~\cite{matuszek2012learning,artzi2013weakly}, generating mathematical statements~\cite{wang2020exploration, sun2019neuralsemantic}, and interacting with time series medical data~\cite{chen2019context}. 

Semantic parsing works typically generate or map a natural language utterance into an unambiguous meaning representation (also referred to as logical forms or semantic representations). The representations need to be executable in an environment or a specific context. The representations typically follow a specified grammar (also referred to as a logical formalism). Grammar is used to identify valid meaning representations. Grammar  can interact with a model which produces a distribution over meaning representations, to ensure that all meaning representations in the distribution are valid.

\textbf{Meaning Representations} of natural language utterances for semantic parsing tend to often fall into one of three categories: first order logic (FOL), graphs, and code snippets. First order logic often expresses unambiguous logical  expressions. It consists of variables and functions that only take objects as arguments. First order logic can be augmented with \textit{lambda calculus} to increase its expressiveness~\cite{carpenter1997type}. An illustrative example from Singh et al.~\citeyearpar{singh2020exploring},  ``a man eating" can be represented as $\exists{A}(\exists{B}(man(A) \land eat(B) \land agent(B, A)))$ in first order logic. First order logic has been used in various semantic parsing works to query databases~\cite{zettlemoyer2005learning}, knowledge bases~\cite{berant2013semantic}, Wikipedia tables~\cite{pasupat2015compositional}, parsing conversations and implementing conversational agents~\cite{artzi2011bootstrapping}, and mapping linguistic instructions onto robotic actions~\cite{artzi2013weakly}. Graph-based meaning representations often denote a labeled graph where nodes represents entities or events and edges represent semantic relationships between the nodes. Graphs tend to be easy to understand compared to other meaning representations like code snippets and first order logic. Graph-based meaning representations also have the advantage in that they also provide abstraction and can pivot away from any general syntax. Notable examples of graph-based meaning representations include Semantic Dependency Parsing~\cite{oepen2015semeval,fernandez2020transition}, Abstract Meaning Representation (AMR)~\cite{banarescu2013abstract}, and Universal Conceptual Cognitive Annotation~\cite{abend2013universal}. Code snippets in general purpose programming languages provide a good medium for representing meaning since they are domain specific and easily executable.  More recently there have been various works that focused on converting natural language into equivalent executable code snippet representations in languages such as SQL~\cite{zhong_seq2sql_2017}, Python~\cite{yin2017syntactic}, and Bash~\cite{lin2018nl2bash}. These works will be discussed later on in~\S{\ref{subsec:code_gen}}. Figure~\ref{fig:semantic_parsing} showcases example semantic parsing tasks with an AMR, a logical form, and SQL, a database specific programming language.  

\textbf{Grammar} of meaning representations defines rules to determine whether a representation is valid or not. Grammar also plays a role in determining how a semantic parser expresses its meaning representations and how computationally complex building the meaning representation is. A notable example of strict grammar for meaning representations is \textit{Combinatory Categorical Grammar} (CCG)\footnote{\href{https://yoavartzi.com/tutorial/}{https://yoavartzi.com/tutorial/}}\cite{steedman2000syntactic,artzi2014semantic}. It is efficiently parsable, but still expressive. CCG has a lot of rules and hence it makes it easier to parse since it lessens the amount of possible meaning representations for a given input~\cite{zettlemoyer2007online}.  Kwiatkowski et al.~\citeyearpar{kwiatkowski2011lexical} consider learning a probabilistic model of CCG grammar for semantic parsing. Recent works on code generation from natural language leverage the programming language grammar itself by using an  abstract syntax tree associated with the programming language.

\textbf{Context} of semantic parsing tasks form the bases of mapping natural language to its meaning representation. Often these contexts or environments are executable. Example semantic parsing contexts include knowledge bases~\cite{berant2013semantic}, Wikipedia tables~\cite{pasupat2015compositional}, geography queries~\cite{zelle1996learning}, SQL databases~\cite{zhong_seq2sql_2017}, spreadsheets~\cite{gulwani2012spreadsheet}, and programming languages~\cite{yin_tranx_2018,yin_syntactic_2017}.

\textbf{Models} for semantic parsing relied on rule-based techniques (e.g. \cite{johnson1984commercial,woods1973progress,hendrix1978developing}), symbolic artificial intelligence (e.g. \cite{zelle1996learning,zelle1993learning}), statistical-based techniques (e.g. \cite{zettlemoyer2005learning,tang2001using}), RNN-based sequence-to-sequence (e.g. \cite{chen2019context,jia_data_2016,singh2020exploring,yin2017syntactic,rabinovich_abstract_2017,DBLP:journals/corr/LingGHKSWB16, sun2019neuralsemantic}), and transformer-based architectures (e.g. \cite{kacupaj2021conversational, ferraro2020transformer, shenetal2019multi,sun_treegen_2019,gemmell_relevance_2020, svyatkovskiy2020intellicode,kusupati_natural_nodate}).  In the late 1970's, Hendrix et al.~\citeyearpar{hendrix1978developing} pioneer the task of interfacing with databases using natural language. The rule-based system proposed called LADDER which takes in a restricted set of natural language questions. LADDER accepts questions if they match a preset template, extract field and file names, and then produces a query or a sequence of queries in a language \textit{Data Language} by inserting the relevant file and field names into preset templates. In 1993, Zelle et al.~\citeyearpar{zelle1993learning} introduce a first-order induction algorithm that utilizes symbolic artificial intelligence to learn word classes an semantic relationships between the words to support parsing of sentences and conduct semantic analyses. The algorithm takes in a set of training samples containing sentences paired with 'case representations' (e.g. parts of speech tags). The algorithm utilizes shift-reduce parsing where in each time step either a new word from the sentence is added (shifted) onto a stack or the top two elements on the stack are popped from the stack then merged to form a new element and pushed back into the stack. After the training is complete the shift-reduce parser is introduced to search control heuristics to help the parser become more specialized. In 2005, Zettlemoyer and Collins~\citeyearpar{zettlemoyer2005learning} propose a statistical probabilistic model that learns  given a training set to induce grammar for mapping natural language sentences to lambda calculus. The probabilistic model learns given a sentence in lambda calculus $L$ and a parse tree $T$. The parse tree $T$  is defined as the sequence of transformations needed to derive $L$ from the given natural language sentence $S$ within the constrains of the grammar CCG. The conditional distribution would be $P(L,T|S)$. In 2020, Singh et al.~\citeyearpar{singh2020exploring} propose using an LSTM-based sequence-to-sequence with Bahdanau attention~\cite{bahdanau_neural_2014} to convert natural language utterances into first order logic. The authors report on improving upon the sequence-to-sequence by introducing a mechanism that aligns variables across predicted predicates in first order logic. This mechanism utilizes a classifier at each decoding step of a variable token to predict whether the variable is aligned with any previously decoded tokens or whether it's a new variable. Following the classifier is self-attention-like mechanism where the decoder hidden states are used to estimate alignment between the variable and other previous decoded tokens. Chen and Bunescu~\citeyearpar{chen2019context} design an LSTM-based encoder decoder model to parse natural language or GUI interactions into a database-executable lambda-calculus logical form. The database contains time series data from sensors monitoring type I diabetes patients along with patient-reported information about discrete life events (e.g. stress, sleep, meals, etc. The data is temporal and user interactions are sequential, where the current interaction may dependent on a prior interaction with the system. This is illustrated in the Figure~\ref{fig:calculus} example. Hence, it is crucial for the model to understand coreferential time expressions such as ``then" and temporal relations between entities such as ``after". To enhance the model with such understanding capabilities, the encoder decoder architecture works with a copying mechanism to model context dependency and utilizes 3 attention mechanisms: the first attention mechanism attends over the previous input, the second attends over the previous logical form, and the third attends over the current input. The proposed approach scores 66.9 in exact match accuracy significantly outperforms the baseline of a standard LSTM-based seq2seq model~\cite{bahdanau2017,cho2014learning} which scored 22.2 accuracy on a real-world dataset. Kacupaj et al.~\citeyearpar{kacupaj2021conversational} employ a transformer architecture to generate logical forms in in a question-answering conversational assistant capable of answering complex questions by utilizing a pre-filled knowledge graph. Ferraro et al.~\citeyearpar{ferraro2020transformer} compare the transformer architecture with other statistical and neural semantic parsing systems on the ATIS~\cite{dahl1994expanding} and Geo~\cite{caiyates2013semantic} semantic parsing datasets and find that the transformer architecture outperforms prior strong models in certain settings and achieve competitive results across all experimental settings.


\begin{figure*}[htbp!]
\centering
\begin{subfigure}[h]{0.5\textwidth}
    \centering
    \includegraphics[height=2.5in]{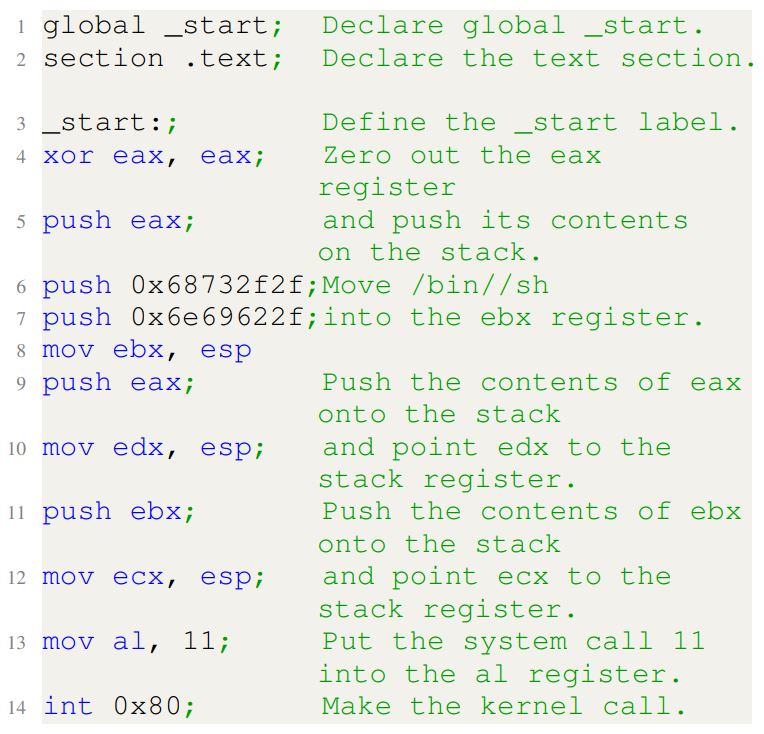}
    \caption{An assembly code generation task. The task is to generate the assembly code that is then compiled into shellcode (small pieces of code used as a payload to exploit software vulnerabilities) using the natural language descriptions on the right. The dataset contains multi-line snippets mapping onto one intent. Lines 4-5, 6-7-8, 9-10, 11-12 are multi-line snippets~\cite{liguori2021shellcode}.}
    \label{fig:assembly}
  \end{subfigure}%
 ~ 
 \begin{subfigure}[h]{0.5\textwidth}
    \centering
    \includegraphics[height=1.75in]{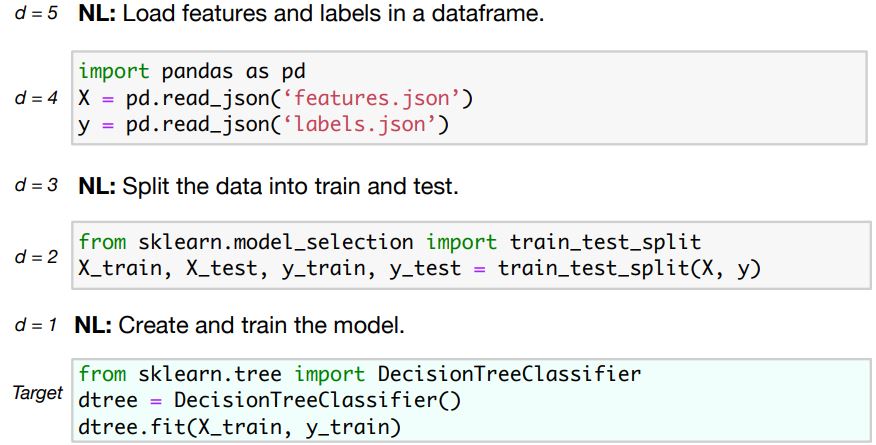}
    \caption{A Python code generation task where models are tasked with generating the blue code cell given the natural language utterance \textit{Create and train the model} along with the previous natural language and code cells in the Jupyter notebook~\cite{agashe2019juice}.}
    \label{fig:python}
  \end{subfigure}\par\medskip

  \caption{Example code generation parsing tasks with various programming languages.}
  \label{fig:codegen_example}
  
\end{figure*}
Semantic parsing is a unique area with the aim of improving natural language understanding (NLU). It is distinguished from natural language generation and machine translation. Both natural language generation and some semantic parsing works use similar deep learning models and techniques such as an LSTM based sequence-to-sequence architecture in the works of Yin and Neubig~\cite{yin2017syntactic}. The objective of semantic parsing involves the prediction or generation of inherently structured representations (not only utterances). These representations need to be executable against a context or an environment. Semantic parsing is related to \textit{Code Generation from Natural Language}, since both predict structured representations of natural language, in this case these structure representations would be code snippets which are executable in an environment and can also be represented as abstract syntax trees. However, semantic parsing also includes other executable representations that are not limited to code such as first order logic and graphs. For other survey papers on semantic parsing c.f.~\cite{kamath_survey_2019, lee2021toward}.


\subsection{Code Generation}
\label{subsec:code_gen}
This section overviews generating a snippet of code given a natural language description.
\subsubsection{Overview}
Code generation is the task of generating an executable snippet given a description in natural language. The executable snippet can be in a particular programming language or domain specific language (DSL) that can be trans-compiled into a programming language.  To illustrate consider the following example generating a Python snippet from an English intent: \textit{check if all elements in list `mylist' are the same $\rightarrow$ len(set(mylist)) == 1.} Code generation from natural language focuses on linking natural language with source code which is something it has in common with other  aforementioned code intelligence tasks code search and semantic parsing. Code generation is distinct from code search. Code search focuses on ranking code snippets and codebases given a natural language query, whilst code generation attempts to generate a meaning representation, an executable code snippet that aligns with a natural language description. With that being said, code generation is closely related to the area of semantic parsing. It can be considered a subfield of semantic parsing that specializes in generating executable code snippets in programming languages an excludes any other meaning representation of natural language. Code generation has been inspired by Natural Language Generation methods including statistical machine translation, RNN-based sequence-to-sequence, and generative transformer models.  Linking codebases to natural language intents or other forms of unstructured input compose face a series of difficult challenges. Firstly, the compositional and abstract nature of code can make pairing text with code difficult. Secondly, natural language is ambiguous in nature while source code is not ambiguous. Code generation can enable users to interface with machines using natural language. For instance, Kuhlmann et al.~\citeyearpar{kuhlmann2004guiding} enable interfacing with robots using natural language. Furthermore, code generation enables users to build code more effectively and efficiently and enhance the overall software engineering process. Code software engineering~
\cite{miltner_fly_2019,xu_ide_2021}, robotics~\cite{kuhlmann2004guiding}, and cyber-security~\cite{you2017semfuzz,liguori2021shellcode, hijax2021,liguori2021evil,liguori2022generate}. Figure~\ref{fig:codegen_example} showcases two examples of code generation tasks.

Recent works in code generation have been used to assist data scientists to perform data visualization and file manipulation~\cite{xu_ide_2021}, generating bash commands~\cite{lin2018nl2bash}, generate exploits~\cite{liguori2021evil,liguori2021shellcode,hijax2021,liguori2022generate}, solve interview-level programming questions~\cite{hendrycks2021measuring}, manipulate data~\cite{zhong2018generating},  and generate code snippets from natural language descriptions in many programming languages. These programming languages include but are not limited to Python~\cite{Xu2020IncorporatingEK,DBLP:journals/corr/LingGHKSWB16,liguori2021evil}, Java~\cite{DBLP:journals/corr/LingGHKSWB16}, SQL~\cite{zhong_seq2sql_2017}, Excel macro commands~\cite{gulwani_nlyze_2014}, Assembly~\cite{liguori2021shellcode,liguori2021evil,liguori2022generate}, and JavaScript~\cite{hijax2021}.

\subsubsection{Datasets}
\label{subsubsec:code_gen_datasets}
This section introduces a comprehensive collection of supervised datasets containing natural language (NL) mapped onto programming language snippets (PL). The snippets can be in general purpose programming languages (e.g. Python, Java) or in a Domain Specific Language (DSL). In table~\ref{tab:code_datasets}, we curate a systematic list (to the best of our knowledge) of English NL-PL datasets. 
\newline
\textbf{Survey Method. }The survey process started by searching through publications on GoogleScholar\footnote{\href{https://scholar.google.com/}{https://scholar.google.com/}} with keywords such as ``Code generation", ``Program synthesis from natural language", ``natural language to code", and ``Conversational Semantic Parsing". We collected highly cited papers and carefully reviewed all of the works that cite them for potentially proposed datasets. For each paper we reviewed, we carefully checked the references for any papers proposing datasets as well. We also searched through PapersWithCode\footnote{\href{https://paperswithcode.com/datasets}{https://paperswithcode.com/datasets}} and Huggingface Datasets~\footnote{\href{https://huggingface.co/datasets?sort=downloads\&search=code}{https://huggingface.co/datasets}} for datasets with the tags ``Code Generation", ``Code Search", and ``Code". Datasets that map other forms of input like images of Graphical User Interfaces~\cite{beltramelli2018pix2code}, input-output examples (without natural language descriptions)~\cite{drosos_wrex_2020}, questions in natural language not descriptions of non-query code (e.g. the neural code search dataset~\cite{li2019neural}), and compiled code~\cite{lacomis2019dire} to source code are excluded from the table. We've also excluded large mined datasets that are not processed into NL-PL pairs like the Code-Clippy dataset~\cite{cooper-2021-code-clippy-data}, CodeNet~\cite{puri2021project}, and The Pile~\cite{gao2020pile}. Multi-lingual datasets were not systematically surveyed.
\newline
\textbf{Automatic Exploit Generation. }Automatic exploit generation is defined as an offensive security technique in which software exploits are automatically generated to explore and test critical vulnerabilities before malicious attackers discover such vulnerabilities~\cite{avgerinos_aeg_2011}. With the goal of offensive security in mind, Liguori et al.~\cite{liguori2021shellcode} share a dataset, Shellcode\_IA32, for shellcode generation from natural language descriptions. Shellcodes are compiled from assembly programs for the 32-bit version of the x86 Intel Architecture and contain a payload that is used in exploiting software vulnerabilities. The Shellcode\_IA32 dataset is later on extended in the EVIL-Decoder~\cite{liguori2021evil} dataset. Shellcodes are often encoded using Python software to evade detection by antivirus programs and then decoded using an assembly program to the victim machine. In an attempt to automate the whole pipeline Liguori et al.~\citeyearpar{liguori2021evil} propose the EVIL dataset composed of two different languages. EVIL-Decoder dataset contains full assembly programs along with their NL descriptions that decode a shellcode on a host machine. The EVIL-Encoder dataset extends the general purpose Python Django dataset~\cite{oda_learning_2015} with 1,114 original exploit-oriented snippets. Frempong et al.~\citeyearpar{hijax2021} curate a synthesized dataset of JavaScript snippets used in cross-site scripting (XSS) which are a common web-based attack that target websites.
\newline
\textbf{Bash. }In an effort to make bash terminal (a Linux operating system command line interface) interactions more accessible, Lin et al.~\citeyearpar{lin2018nl2bash} curate a dataset of bash commands along with their corresponding natural language descriptions. This work helps lay down a foundational dataset and a baseline model in building natural language interfaces to terminals.
\newline
\textbf{General Purpose Programming. }There has been extensive work in curating datasets for general purpose programming languages such as Java, JavaScript, and Python, since general purpose programming languages are abundant in open source code -bases~\cite{husain_codesearchnet_2019}. While some datasets were expert curated~\cite{oda_learning_2015,chen2021evaluating}, and crowd sourced ~\cite{long2016simpler, zavershynskyi2018naps, kulal2019spoc, austin2021program,huang-etal-2021-cosqa}, extensive datasets with NL-PL pairs were mined from open source code sources such as Github~\cite{allamanis2016convolutional,barone2017parallel,iyer2018mapping,alon2018code2seq,hu2018deep, husain_codesearchnet_2019, agashe2019juice,clement2020pymt5, hu2020deep, bahrami2021pytorrent}, coding competition websites~\cite{Caballero_Description2Code_Dataset_2016,alphacode}, Sourcerer~\cite{lopes2010data,leclair2019neural}, IFTTT~\cite{quirk_language_2015} (a website that allows users to create simple programs using triggers and actions), and StackOverflow~\cite{iyer_summarizing_2016,yin2018mining,yao2018staqc,orlanski2021reading}. CoDesc~\cite{hasan2021codesc} combines multiple mined Java datasets into one bigger dataset after removing data-related noise. Hendrycks et al.~\citeyearpar{hendrycks2021measuring} propose the Automated Programming Progress Standard, a dataset curated from specifially mining websites where programmers share programming problems with each other such as CodeForces, Codewars, AtCoder, and Kattis. APPS contains Python programming problems in various difficulties (Introductory, Interview, and Competition levels), a natural language description, and unit tests for each problem. Some datasets contain synthetically generated NL descriptions such as the Hearthstone and Magic The Gathering trading card game datasets, were the NL-descriptions were automatically populated fields describing the card~\cite{DBLP:journals/corr/LingGHKSWB16}. Akinbou et al.~\citeyearpar{akinobu2021_coding_assistance} propose using back-translation on Python code collected from Aizu Online Judge (AOJ)\footnote{\href{https://judge.u-aizu.ac.jp/onlinejudge/}{https://judge.u-aizu.ac.jp/onlinejudge/}} using a trans-compiler. The trans-compiler outputs natural language descriptions, in Japanese, from a Python code input. The Python snippet is converted into an abstract syntax tree and then using pre-defined transformation rules, the abstract syntax tree is transformed into a natural language description. Hasan et al.~\citeyearpar{hasan2021text2app} propose text2app, a dataset containing descriptions of Android app specifications along with a Simplified App Representation (SAR) codebase. SAR is a domain specific language that is then compiled into a Java Android app. Text2App used data BERT-masking for data augmentation to curate NL descriptions of apps from a small amount of crowd-sourced app descriptions.
\newline
\textbf{Turducken-style Code Generation. }Turducken-style code snippets include one language embedded in another. Often-times a general purpose programming language like Python being embedded with another server-related programming language like SQL. In an effort to benchmark code generation systems on Turducken-style code, Liang et al.~\citeyearpar{liang2021lyra} propose the Lyra dataset which contains natural language intents in both English and Chinese languages mapped onto Python code with embedded SQL snippets.
\newline
\textbf{Conversational Semantic Parsing. }Semantic parsing is a task in which an utterance is converted into an executable form (\S\ref{subsection:semantic_parsing}). Semantic Parsing works often focus on isolated utterances. Conversational Semantic Parsing takes that a step further, where the context is a conversation. A crucial part of a dialogue system is being able to track the user's goal through out the conversation, this task is called Dialogue State Tracking (DST). In conversational semantic parsing, researchers have formulated the user's goal into an SQL query~\cite{yu2019cosql}, a tree-like representation using a domain specific language~\cite{gupta-etal-2018-semantic-parsing,cheng-etal-2020-conversational,aghajanyan-etal-2020-conversational}, and a program in a domain specific language that extends a dataflow graph~\cite{andreas2020_taskoriented_dataflow}. These representations are executable by a dialogue system to interface with a data source. People tend to explore databases by asking multiple related questions~\cite{iyyer2017search}, which require systems to be able to process conversational data requests, clarify ambiguous questions, and process user utterances that can not be mapped onto queries~\cite{yu2019cosql}. To address this, Yu et al.~\citeyearpar{yu2019cosql} release the Conversational Text-to-SQL (CoSQL) corpus, to build database querying dialogue systems. Human utterances in dialogue sometimes contain nested requests. An example from \cite{gupta-etal-2018-semantic-parsing} illustrates this well: ``Get me driving directions to the Eagles game" it is composed of two requests \texttt{GET\_DIRECTIONS} and \texttt{GET\_EVENT}. Linear representations do not allow for such compositionality. To tackle this phenomena, both the TreeDST~\cite{cheng-etal-2020-conversational} and the TOP datasets~\cite{gupta-etal-2018-semantic-parsing}, use a compositional (tree-like) forms.  SB-TOP builds on the TOP dataset to include a decoupled representation that can represent co-reference and context carry over~\cite{aghajanyan-etal-2020-conversational}.
\newline
\textbf{Database Querying. }Creating natural language interfaces to interacting with data sources such as databases has been a long standing research~\cite{price-1990-evaluation,dahl1994expanding}. To this end, various hand-curated datasets containing NL-descriptions and data-related questions mapping onto a domain specific language (such as a logical form) trans-compilable into a query in various domains such as Airline Travel in English~\cite{price-1990-evaluation,dahl1994expanding} and other various languages~\cite{upadhyay2018almost,xu2020end}, Job posts~\cite{tang2001using}, open-domain question answering~\cite{caiyates2013semantic, berant2013semantic,wang2015building,yihetal2016value}. We also observe an array of that focus on generating SQL queries from natural language. Some of these datasets are synthetic~\cite{zhong_seq2sql_2017}, mined from StackOverflow~\cite{yao2018staqc,hazoom2021text} and Github~\cite{yao2018staqc}, and human-curated~\cite{data-restaurants-logic,data-restaurants-original,data-restaurants,data-academic, iyer2017learning,yuetal2018spider,data-sql-imdb-yelp,finegan-dollak-etal-2018-improving,yu2019sparc}.
\newline
\textbf{Map Question-Answering. }Maps (e.g. Google Maps) contain data on entities restaurants, landmarks, geographic landmarks like rivers and mountains. This data is often stored in a database. A specific-type of text-to-query task deals with querying map databases. In an effort to mainstream interfacing through natural language with geographic data in map-related databases, Zelle and Mooney ~\citeyearpar{zelle1996learning} formulated the dataset GeoQuery contain questions about the United States geography in natural language and corresponding them to Prolog programs to answer them. GeoQuery was later adapted to SQL~\cite{iyer2017learning,finegan-dollak-etal-2018-improving}. Another notable dataset that contains more complex natural language questions and entities such as restaurants, museums, and hotels is NLMAPS~\cite{haasriezler2016corpus,lawrence2016NLmaps}. NLMAPS contain manually curated human questions that can be run against a worldwide map data from OpenStreetMap. Natural language intents were later on expanded using synthetic methods in NLMAPS-V2~\cite{lawrence2018a}.
\newline
\textbf{Data Manipulation. }Datases were curated to with the aim of automatically generated codes that manipulate data. Gulwani et al~\citeyearpar{gulwani_nlyze_2014} curated a dataset that maps English utterances onto Excel macro-commands. Other datasets focus on generating a regular expression (Regex)~\cite{kushmanbarzilay2013using,locascioetal2016neural,zhong2018generating} given a natural language description.





\renewcommand{\tabcolsep}{4pt}

\onecolumn


\begin{small}
\begin{landscape}
\begin{longtable}{@{}llcccccccp{3.6cm}@{}
>{\columncolor[HTML]{FFFFFF}}c 
>{\columncolor[HTML]{FFFFFF}}c 
>{\columncolor[HTML]{FFFFFF}}c 
>{\columncolor[HTML]{FFFFFF}}c 
>{\columncolor[HTML]{FFFFFF}}c 
>{\columncolor[HTML]{FFFFFF}}c 
>{\columncolor[HTML]{FFFFFF}}c 
>{\columncolor[HTML]{FFFFFF}}c 
>{\columncolor[HTML]{FFFFFF}}c 
>{\columncolor[HTML]{FFFFFF}}l @{}}
\toprule
\textbf{Dataset Name} &
\textbf{\begin{tabular}[l]{@{}l@{}}Programming \\ Language \\ (PL)\end{tabular}} &
  \textbf{\#NL-PL pairs} &
  \textbf{\#NL tokens} &
  \textbf{\#PL tokens} &
  \textbf{\begin{tabular}[c]{@{}c@{}}\#Avg tokens \\ per NL intent\end{tabular}} &
  \textbf{\begin{tabular}[c]{@{}c@{}}\#Avg tokens per \\ code snippet\end{tabular}} &
  \textbf{\begin{tabular}[c]{@{}c@{}}Data \\ Collection \\ (NL) \end{tabular}} &
  \textbf{Public} &
  \textbf{Reference} \\* \midrule
\endhead
\bottomrule
\endfoot
\endlastfoot
\multicolumn{10}{c}{\cellcolor[HTML]{EFEFEF}\textbf{Automatic Exploit Generation}} \\
Shellcode\_IA32 &
  Assembly &
  3,202.00 &
  1,490.00 &
  1,238.00 &
  9.24 &
  4.40 &
  EC &
  \checkmark   &
  \cite{liguori2021shellcode} \\
  
EVIL-Encoder &
  Python &
 15,540 &
  10,605 &
  9,511 &
  14.90 &
  11.90 &
  EC &
  \checkmark   &
  \cite{liguori2021evil} \\
 EVIL-Decoder &
  Assembly &
  3,715 &
  1,924 &
  1,657 &
  9.53 &
  4.75 &
  EC &
  \checkmark   &
  \cite{liguori2021evil} \\
HIJAX &
  JavaScript &
  100,000 &
  - &
  - &
  - &
  - &
  Synth &
  $\times$   &
  \cite{hijax2021} \\
\multicolumn{10}{c}{\cellcolor[HTML]{EFEFEF}\textbf{Bash}} \\
NL2Bash &
  Bash &
  9,305 &
  7,790 &
  6,234 &
  11.70 &
  7.70 &
  EC &
  \checkmark   &
  \cite{lin2018nl2bash} \\
\multicolumn{10}{c}{\cellcolor[HTML]{EFEFEF}\textbf{General Purpose Programming}} \\
IFTTT &
  DSL &
  86,960 &
  593,123 &
  1,881,451 &
  6.82 &
  21.64 &
  Mined &
  $\times$  &
  \cite{quirk_language_2015} \\
\begin{tabular}[c]{@{}c@{}}Code-Docstring Corpus\end{tabular} &
  Python &
  150,370 &
  5,789,741 &
  12,601,929 &
  38.50 &
  83.81 &
  Mined &
  \checkmark   &
  \cite{barone2017parallel} \\
C\#2NL &
  C\# &
  66,015 &
  24,857 &
  91,156 &
  12.00 &
  38.00 &
  Mined &
  \checkmark   &
  \cite{iyer_summarizing_2016} \\
Django &
  Python &
  18,805 &
  - &
  - &
  14.30 &
  - &
  EC &
  \checkmark   &
  \cite{oda_learning_2015} \\
CoNaLa &
  Python &
  \textbf{2,879*} &
  - &
  - &
  - &
  - &
  EC &
  \checkmark   &
  \cite{yin2018mining} \\
CoNaLa Mined &
  Python &
  593,837 &
  - &
  - &
  11.41 &
  28.70 &
  Mined &
  \checkmark   &
  \cite{yin2018mining} \\
CoNaLa-Ext &
  Python &
  596,711 &
  - &
  - &
  11.41 &
  28.70 &
  Mined &
  \checkmark   &
  \cite{orlanski2021reading} \\
\begin{tabular}[c]{@{}c@{}}Magic The Gathering\end{tabular} &
  Java &
  13,297 &
  - &
  - &
  21.00 &
  1,080.00 &
  Synth &
  \checkmark   &
  \cite{DBLP:journals/corr/LingGHKSWB16} \\
CONCODE &
  Java &
  104,000 &
   &
   &
  13.73 &
  26.30 &
  Mined &
  \checkmark   &
  \cite{iyer2018mapping} \\
HearthStone &
  Python &
  665 &
  - &
  - &
  7.00 &
  352.00 &
  Synth &
  \checkmark   &
  \cite{DBLP:journals/corr/LingGHKSWB16} \\
JuICe &
  Python (Jupyter) &
  1,521,774 &
  862,269 &
  1,006,402 &
  39.78 &
  38.75 &
  Mined &
  \checkmark   &
  \cite{agashe2019juice} \\
Text2App &
  SAR (DSL)* &
  50,000 &
  - &
  - &
  - &
  - &
  Synth &
  \checkmark   &
  \cite{hasan2021text2app} \\
DeepCom &
  Java &
  588,108 &
  - &
  794,711 &
  8.86 &
  99.94 &
  Mined &
  \checkmark   &
  \cite{hu2018deep,hu2020deep} \\
FunCom &
  Java &
  2,100,000 &
  - &
  - &
  - &
  - &
  Mined &
  \checkmark   &
  \cite{lopes2010data,leclair2019neural} \\
PyMT5 &
  Python &
  7,700,000 &
  - &
  - &
  - &
  - &
  Mined &
  $\times$  &
  \cite{clement2020pymt5} \\
CodeSearchNet & 
\begin{tabular}[c]{@{}l@{}}Python, Java, Go, \\ JavaScript,\\ Ruby, and PHP\end{tabular} &
  2,326,976 &
  - &
  - &
  - &
  - &
  Mined &
  \checkmark   &
  \cite{husain_codesearchnet_2019} \\
Java-small &
  Java &
  665,115 &
  - &
  - &
  3.00 &
  60.00 &
  Mined &
  \checkmark   &
  \cite{allamanis2016convolutional,alon2018code2seq} \\
Java-med &
  Java &
  3,004,536 &
  - &
  - &
  3.00 &
  63.00 &
  Mined &
  \checkmark   &
  \cite{alon2018code2seq} \\
Java-large &
  Java &
  15,344,512 &
  - &
  - &
  3.00 &
  65.00 &
  Mined &
  \checkmark   &
  \cite{alon2018code2seq} \\
StaQC-Python &
  Python &
  147,546 &
  17,635 &
  137,123 &
  9.00 &
  86.00 &
  Mined &
  \checkmark   &
  \cite{yao2018staqc} \\
Euler &
  Python &
  589 &
  - &
  - &
  - &
  - &
  EC &
  \checkmark   &
  \cite{oda_learning_2015} \\
AOJ &
  Python &
  89,862 &
  - &
  - &
  - &
  - &
  Synth-BT &
  $\times$  &
  \cite{akinobu2021_coding_assistance} \\
NAPS &
  DSL &
  2,716 &
  - &
  - &
  - &
  - &
  CC &
  \checkmark   &
  \cite{zavershynskyi2018naps} \\
APPS &
  Python &
  10,000 &
  - &
  - &
  - &
  - &
  Mined+revision &
  \checkmark   &
  \cite{hendrycks2021measuring} \\
  CodeContests &
  Python, C++, Java &
  13,610 &
  - &
  - &
  - &
  - &
  Mined &
  \checkmark   &
  \cite{alphacode} \\
  Description2Code &
  Python, C++ &
  7,764 &
  - &
  - &
  - &
  - &
  Mined &
  \checkmark   &
  \cite{Caballero_Description2Code_Dataset_2016} \\
SCONE-Scene &
  DSL &
  4,402 &
  - &
  - &
  56.20 &
  - &
  CC &
  \checkmark   &
  \cite{long2016simpler} \\
SCONE-Alchemy &
  DSL &
  4,560 &
  - &
  - &
  39.90 &
  - &
  CC &
  \checkmark   &
  \cite{long2016simpler} \\
SCONE-Tangrams &
  DSL &
  4,989 &
  - &
  - &
  27.20 &
  - &
  CC &
  \checkmark   &
  \cite{long2016simpler} \\
CoDesc &
  Java &
  4,200,000 &
  813,078 &
  1,128,909 &
  21.04 &
  77.97 &
  Mined &
  \checkmark   &
  \cite{hasan2021codesc} \\
HumanEval &
  Python &
  164 &
  - &
  - &
  - &
  - &
  EC &
  \checkmark   &
  \cite{chen2021evaluating} \\
MBPPS &
  Python &
  974 &
  - &
  - &
  - &
  - &
  CC &
  \checkmark   &
  \cite{austin2021program} \\
Math-QA &
  Python &
  23,914 &
  - &
  - &
  - &
  - &
  CC &
  \checkmark   &
  \cite{amini-etal-2019-mathqa,austin2021program} \\
PyTorrent &
  Python &
  13,825,647 &
  - &
  - &
  - &
  - &
  Mined &
  \checkmark   &
  \cite{bahrami2021pytorrent} \\
SPoC &
  C++ &
  18,356 &
  - &
  - &
  - &
  - &
  CC &
  \checkmark   &
  \cite{kulal2019spoc} \\
CoSQA &
  Python &
  20,604 &
  6,784 &
  28,254 &
  6.60 &
  71.51 &
  CC &
  \checkmark   &
  \cite{huang-etal-2021-cosqa} \\
\multicolumn{10}{c}{\cellcolor[HTML]{EFEFEF}\textbf{Turducken-style Code Generation}} \\
Lyra-English &
  Python + SQL &
  2,000 &
  - &
  - &
  57.71 &
  44.24 &
  EC &
  \checkmark   &
  \cite{liang2021lyra} \\
Lyra-Chinese &
  Python + SQL &
  2,000 &
  - &
  - &
  70.46 &
  44.24 &
  EC &
  \checkmark   &
  \cite{liang2021lyra} \\
\multicolumn{10}{c}{\cellcolor[HTML]{EFEFEF}\textbf{Conversational Semantic Parsing}} \\
CoSQL &
  SQL &
  11,039 &
  9,585 &
  - &
  11.21 &
  - &
  CC &
  \checkmark   &
  \cite{yu2019cosql} \\
TOP &
  DSL &
  44,000 &
  - &
  - &
  8.93 &
  - &
  CC &
  \checkmark   &
  \cite{gupta-etal-2018-semantic-parsing} \\
SB-TOP &
  DSL &
  60,000 &
  - &
  - &
  - &
  - &
  CC &
  \checkmark   &
  \cite{aghajanyan-etal-2020-conversational} \\
TreeDST &
  DSL &
  167,507 &
  - &
  - &
  7.59 &
  - &
  CC &
  \checkmark   &
  \cite{cheng-etal-2020-conversational} \\
 SMCalFlow &
  DSL &
  155,923 &
  17,397 &
  338 &
  8 &
  40 &
  CC &
  \checkmark   &
  \cite{andreas2020_taskoriented_dataflow} \\

\multicolumn{10}{c}{\cellcolor[HTML]{EFEFEF}\textbf{Database Querying}} \\
ATIS &
  DSL &
  5,410 &
  936 &
  176 &
  11.10 &
  28.10 &
  CC &
  \checkmark   &
  \cite{price-1990-evaluation,dahl1994expanding} \\

Multi-ATIS &
  DSL &
  3,846 &
  - &
  - &
  - &
  - &
  CC &
  \checkmark   &
  \cite{upadhyay2018almost} \\
  
Multi-ATIS++ &
  DSL &
  44,943 &
  - &
  - &
  - &
  - &
  CC &
  \checkmark   &
  \cite{xu2020end} \\

Freebase917 &
  DSL &
  917 &
  - &
  - &
  - &
  - &
  EC &
  \checkmark   &
  \cite{caiyates2013semantic, berant2013semantic} \\

Jobs640 &
  DSL &
  640 &
  391 &
  58 &
  9.80 &
  22.90 &
  EC &
  \checkmark   &
  \cite{tang2001using} \\
WebQSP &
  DSL &
  4,737 &
  - &
  - &
  - &
  - &
  EC &
  \checkmark   &
  \cite{yihetal2016value} \\
WikiSQL &
  SQL &
  80,654 &
  - &
  - &
  - &
  - &
  Synth &
  \checkmark   &
  \cite{zhong_seq2sql_2017} \\
StaQC-SQL &
  SQL &
  119,519 &
  9,920 &
  21,413 &
  9.00 &
  60.00 &
  Mined &
  \checkmark   &
  \cite{yao2018staqc} \\
SQL2NL &
  SQL &
  32,337 &
  10,086 &
  1,287 &
  9.00 &
  46.00 &
  Mined &
  \checkmark   &
  \cite{iyer_summarizing_2016} \\
  
SParC &
  SQL &
  12,726 &
  3,794 &
  3,794 &
  8.10 &
  - &
  EC &
  \checkmark   &
  \cite{yu2019sparc} \\
  
SPIDER &
  SQL &
  10,181 &
  - &
  - &
  13.00 &
  21 &
  EC &
  \checkmark   &
  \cite{yuetal2018spider} \\
Restaurants &
  SQL &
  378 &
  - &
  - &
  - &
  - &
  EC &
  \checkmark   &
  \cite{data-restaurants-logic,data-restaurants-original,data-restaurants} \\
Scholar &
  SQL &
  817 &
  - &
  - &
  - &
  - &
  EC &
  \checkmark   &
  \cite{iyer2017learning} \\
Yelp &
  SQL &
  128 &
  - &
  - &
  - &
  - &
  EC &
  \checkmark   &
  \cite{data-sql-imdb-yelp} \\
IMDB &
  SQL &
  131 &
  - &
  - &
  - &
  - &
  EC &
  \checkmark   &
  \cite{data-sql-imdb-yelp} \\
Advising &
  SQL &
  4,570 &
  - &
  - &
  - &
  - &
  EC &
  \checkmark   &
  \cite{finegan-dollak-etal-2018-improving} \\
Academic &
  SQL &
  196 &
  - &
  - &
  - &
  - &
  EC &
  \checkmark   &
  \cite{data-academic} \\
Overnight &
  DSL &
  12,602 &
  - &
  - &
  - &
  - &
  CC &
  \checkmark   &
  \cite{wang2015building} \\
SEDE &
  SQL &
  12,023 &
  - &
  - &
  - &
  - &
  Mined &
  \checkmark   &
  \cite{hazoom2021text} \\

\multicolumn{10}{c}{\cellcolor[HTML]{EFEFEF}\textbf{Map Question-Answering}} \\
NLMAPS-V1 &
  DSL &
  2,380 &
  1,014 &
  - &
  10.90 &
  16.00 &
  EC &
  \checkmark   &
  \cite{haasriezler2016corpus,lawrence2016NLmaps} \\
NLMAPS-V2 &
  DSL &
  202,088 &
  8,710 &
  - &
  7.06 &
  - &
  Synth &
  \checkmark   &
  \cite{lawrence2018a} \\
  
GeoQuery &
  Prolog, SQL &
  880 &
  284 &
  60 &
  7.60 &
  19.10 &
  EC &
  \checkmark   &
  \cite{zelle1996learning,iyer2017learning,finegan-dollak-etal-2018-improving} \\
  
\multicolumn{10}{c}{\cellcolor[HTML]{EFEFEF}\textbf{Data Manipulation}} \\
NLyze-Data &
  DSL (Excel) &
  3,570 &
  - &
  - &
  - &
  - &
  CC &
  $\times$  &
  \cite{gulwani_nlyze_2014} \\
RegexLib &
  Regex &
  3,619 &
  13,491 &
  179 &
  36.40 &
  58.80 &
  Mined &
  $\times$  &
  \cite{zhong2018generating} \\
NL2RX &
  Regex &
  10,000 &
  560 &
  45 &
  10.60 &
  26.00 &
  Synth &
  \checkmark   &
  \cite{locascioetal2016neural} \\
NL2RX-KB13 &
  Regex &
  824 &
  715 &
  85 &
  7.10 &
  19.00 &
  CC &
  \checkmark   &
  \cite{kushmanbarzilay2013using} \\ \bottomrule
\caption{Survey of datasets containing natural-language-programming language (NL-PL) pairs that are usable for the tasks of code generation and semantic parsing. We also include domain specific languages (DSL) if they can be trans-compiled into a programming language. EC stands for expert curated, CC, stands for crowdsourced, Synth stands for synthesized, Synth-BT sands for synthesized with back-translation. The authors will add footnote to table for datasets reporting on total number of tokens vs total number of unique tokens.}
\label{tab:code_datasets}\\
\end{longtable}
\end{landscape}
\end{small}

\twocolumn
\subsubsection{Evaluation}
Code Generation systems are generally evaluated using exact match accuracy. Exact match accuracy refers to the ratio of model-generates snippets that exactly match the ground-truth snippets. Another common metric used for evaluating code generation systems is token level BLEU~\cite{papineni2002bleu}. It is commonly used to evaluate machine translation systems. BLEU is used to evaluate code generation systems since many prior works in code generation formulated the problem as a machine translation problem of translating English to code snippets (e.g.~\cite{liguori2021shellcode}). Both exact match and averaged token level BLEU scores have been extensively used in evaluating code generation models~\cite{liguori2021shellcode,liguori2021evil,oda2015learning,DBLP:journals/corr/LingGHKSWB16,gemmell_relevance_2020}. It is becoming increasingly important to note the drawback of using BLEU to evaluate code generation systems. Recent studies have shown low correlation between BLEU score and program correctness~\cite{austin2021program,chen2021evaluating,hendrycks2021measuring}, i.e. where the generated program logically matches the natural language description and passes all the unit tests. BLEU score increases when there is significant token overlap between the generated code and the ground-truth. The low correlation  can be attributed to the abundance of variable and method identifiers in programming languages, which if the model predicts the identifier tokens correctly, it would result in a higher BLEU score. However, even though there may be signification token overlap the generated code may not compile (be syntactically incorrect), token differences between the ground truth and generated code may cause the generate code to follow a distinct logic. Furthermore, there may be more than one correct code snippet that accomplishes the task described in natural language. Both token-level BLEU and exact match accuracy penalize models for generating working code snippets that are different from the ground truth, even if they are correct. 
To remedy this, recent works in code generation such as APPS~\cite{hendrycks2021measuring}, HumanEval~\cite{chen2021evaluating}, and EVIL~\cite{liguori2021evil} focus on evaluating code generation systems by formulating metrics focused on evaluating the semantics of the generated snippets. Liguori et al.~\citeyearpar{liguori2021evil} define two new metrics of syntactic and semantic correctness. Syntactic correctness is the ratio of snippets that are compilable. This is achieved by compiling the generated snippets with a compiler and recording the ratio of generated snippets that do compile. For semantic correctness, domain experts were asked to evaluate whether generated code snippets semantically express the natural language description. The APPS dataset evaluates generated snippets based on the percentage of unit tests the generated snippets pass, this metric is called the Test Case Average~\cite{hendrycks2021measuring}. Additionally, APPS evaluate generated codes on their ability to pass all unit tests for a particular problem, this metric is called Strict Accuracy. Similarly in the HumanEval dataset, the authors introduce a $pass@k$ metric, in which $k$ code snippets are sampled from the model. $pass@k$ would be the ratio of problems in which any of the $k$ samples pass the problem's corresponding unit tests. Another notable metric proposed for evaluating code generation systems is CodeBLEU~\cite{ren2020codebleu}. CodeBLEU\footnote{\href{https://github.com/microsoft/CodeXGLUE/tree/main/Code-Code/code-to-code-trans/evaluator/CodeBLEU}{https://github.com/microsoft/CodeXGLUE/tree/main/Code-Code/code-to-code-trans/evaluator/CodeBLEU}}, combines the n-gram match inspired from the BLEU score, along with weighed N-gram match, abstract syntax tree match, and semantic data-flow match into one metric. The drawback of CodeBLEU is it needs to have language specific AST-parsers implemented which may not be available to low-resource programming languages like Haskell and Assembly yet. Current CodeBLEU supports 6 programming languages: Java, JavaScript, C\#, PHP, GO, Python, and Ruby.

\subsubsection{Methods}
\label{subsubsec:code_gen_methods}
\begin{figure*}[h!]
  \includegraphics[width=\linewidth]{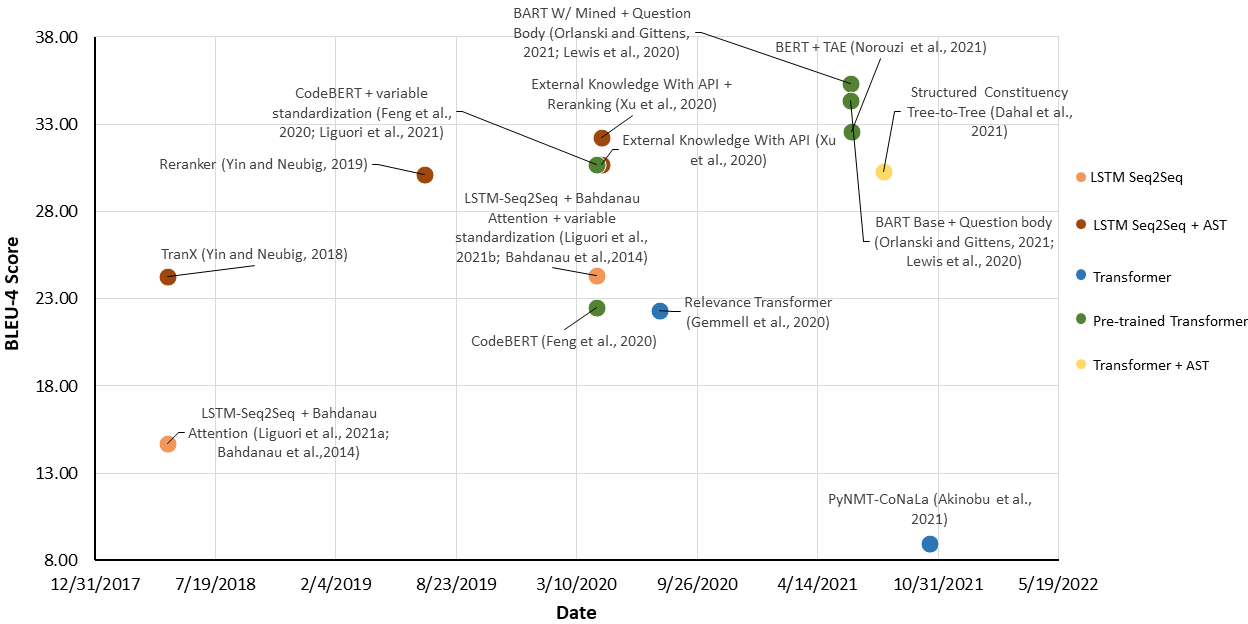}
  \caption{Comprehensive review of all code generation models evaluated on the CoNaLa dataset~\cite{yin2018mining} across time. Models are scored using BLEU-4~\cite{papineni2002bleu} and use one of the following deep learning architectures: LSTM-based sequence to sequence, LSTM-based sequence to sequence models that encode Abstract Syntax Trees (ASTs), Transformer-based models, Transformer models loaded with pretrained weights, and Transformer models that encode Abstract Syntax Trees. We note that the LSTM-Seq2Seq with Bahdanau attention~\cite{bahdanau_neural_2014} and CodeBERT~\cite{feng_codebert_2020} experiments were conducted by the authors of this paper. The method of variable standardization is described in~\cite{liguori2021evil,liguori2022generate}.}

  \label{fig:conala}
\end{figure*}

Here we introduce  different deep learning architectures used for the task of code generation. We categorize the approaches into LSTM-based sequence to sequence, LSTM-based sequence to sequence with abstract syntax trees, transformers, pretrained transformers, and transformers with abstract syntax trees. We survey works in each of these approaches and describe them in detail. We focus on the CoNaLa dataset~\cite{yin2018mining} as a case study in Figure~\ref{fig:conala} and overview all the works that have benchmarked their systems on this dataset.

\textbf{LSTM-based Seq2Seq.}  Ling et al.~\citeyearpar{DBLP:journals/corr/LingGHKSWB16} LSTM based sequence-to-sequence model called the Latent Predictor Network (LPN) has been proposed for general-purpose programming languages (Python and Java) it includes a copying mechanism similar to ~\cite{gu2016incorporating} to copy over identifier names. A weakness in that approach is that it does not consider code syntax in its processing and output. Dong et al.~\citeyearpar{dong2018coarse} developed \textit{Coarse-to-Fine Decoding}, a methodology for semantic parsing and code generation simultaneously. Coarse-to-Fine Decoding essentially uses two neural networks. The first neural network is used to parse the input into a rough sketch. The rough sketch preserves the original meaning of the input while removing lower-level details. This sketch and the original input is fed into the second neural network which then produces the output. This method is better than simply parsing the input through a single neural network. The results show that this method increases performance when compared to using one neural network to parse the input to code. Additionally, the Coarse-to-Fine Decoding can be used for a variety of parsing tasks making it a fairly flexible method. Lin et al.~\citeyearpar{lin2018nl2bash} use an LSTM-based Seq2Seq and CopyNet~\cite{gu2016incorporating} to generate linux \texttt{bash} commands from natural language descriptions. Zhong et al.~\citeyearpar{zhong_seq2sql_2017} develop an LSTM-based sequence to sequence model adapted from Dong and Lapta~\citeyearpar{dong-lapata-2016-language}  with a pointer network~\cite{vinyals_2015_pointer} containing augmented inputs specific to the SQL language. Chen et al.~\citeyearpar{chen-2021-plotcoder} propose PlotCoder, an LSTM-based Seq2Seq with a copying mechanism trained on a portion on the JuICe dataset~\cite{agashe2019juice} trained to generate plot code snippets in Python Liguori et al.~\citeyearpar{liguori2021shellcode} use a standard LSTM-based sequence to sequence model~\cite{bahdanau_neural_2014} to generate assembly snippets from natural language intents that are then compiled into shellcodes used in exploiting vulnerabilities. Frempong et al.~\citeyearpar{hijax2021} generate JavaScript XSS exploits from natural language intents. Weaknesses of LSTM-based models include generating syntactically incorrect code, however given enough training samples or on syntactically simple languages like Assembly, LSTM-based models tend to generate syntactically correct code with good accuracy~\cite{liguori2021evil}.

\textbf{LSTM-based Seq2Seq with Abstract Syntax Trees.} To remedy generated syntactically incorrect codes, Yin and Neubig~\citeyearpar{yin2017syntactic} propose a novel LSTM-based architecture that generates an Abstract Syntax Tree (AST) given a natural language description and a grammar model.  Their system includes a set of production rules for python’s abstract grammar, a probabilistic grammar model, and finally a Recurrent Neural Network with the dimensions of 256 by 50 with a beam size of 15 for the decoder. The results in the model mentioned above scored 16.2 exact match accuracy and a 75.8 BLEU-4 on the Hearthstone dataset~\cite{DBLP:journals/corr/LingGHKSWB16}, scoring 11.7 accuracy above the state-of-the-art at the time the Latent Predictor Network model. The proposed system’s weaknesses include mismatching parameter names when defining a function, omitting or adding default values of parameters when defining a function, failing to copy a variable name into the correct position, and lastly the generated code partially implemented the required functionality. Rabinovich et al.~\citeyearpar{rabinovich_abstract_2017} leveraged the Abstract Syntax Description Language (ASDL) framework for semantic parsing and a Long Short-term Memory network (LSTM) for code generation. These trees include two types of modules–composite types (function and class definitions, return statements, etc) and primitive types (integers, identifiers). Composite modules use LSTM to determine the appropriate constructor module which details how the node should expand. These constructor modules use LSTM to determine what constructor field modules are needed. TranX~\cite{yin_tranx_2018} uses Abstract Syntax Trees as an intermediary representation (\textit{interlingua}) to generate a  programming language snippet from a natural language statement. TranX achieved state of the art on the CoNaLa dataset~\cite{yin2018mining} at the time scoring 24.3 BLEU-4~\cite{yin2019reranking}. Yin and Neubig~\citeyearpar{yin2019reranking} then incorporate a hypothesis reranker to rerank the outputs of TranX and see a 5.7\% increase in BLEU on the CoNaLa dataset setting state of the art at 30.11 BLEU-4. Xu et al. \citeyearpar{Xu2020IncorporatingEK} then explore incorporating external knowledge from API documentation and mined NL-PL pairs from StackOverflow to the TranX model~\cite{yin_tranx_2018} with the hypothesis reranking proposed in~\cite{yin2019reranking}. This approach improves upon the reranker model by ~2.2\% scoring 32.26 BLEU. We note that LSTM methods with ASTs significantly outperform the standard LSTM sequence to sequence scoring around 14.72 BLEU on the CoNaLa dataset. Another notable work, although not used in code generation to our knowledge, is Code2Seq~\cite{alon2018code2seq}. Code2seq encodes the source code tokens along with the AST paths simultaneously and has seen good success in the area of Code Documentation Generation \& Summarization (c.f.\S\ref{subsec:comment_gen}).

Incorporating structural knowledge from ASTs into a neural deep learning model ensures correct syntax of generated code all the time, it also improves model generalizability to NL intents~\cite{yin2017syntactic}. The drawback is incorporating ASTs will take time since grammar rules normally need to be specified.

\textbf{Transformer Models. }Bonthu et al.~\citeyearpar{bonthu2021text2pycode} use a standard transformer architecture~\cite{vaswani_attention_2017} to generate Python source code from natural language intents and report BLEU score of 32.4 and Rouge-L of 82.1 on a custom curated dataset. Liang et al.~\citeyearpar{liang2021lyra} set a baseline for a turducken-style code generation task where SQL is embedded with Python. The baseline used is a transformer model and it achieves ~48-49 BLEU and ~18\%-21\% executable codes when generating codes from the English and Chinese natural language intents respectively. Akinbou et al.~\citeyearpar{akinobu2021_coding_assistance} use a standard transformer architecture and achieve 8.97 BLEU on the CoNaLa dataset with a 42\% syntax error rate. Gemmell et al.~\citeyearpar{gemmell_relevance_2020} utilize a transformer architecture to translate natural language intents to executable snippet representations in Python. The authors further improve upon the baseline transformer model by introducing pseudo-relevance feedback during the decoding process. During decoding process top $k$ documents relevant to the input are retrieved and a set of common tokens from the documents are emphasized. The authors additionally use copying mechanisms inspired by prior work in pointer networks~\cite{see2017get} to assist with tokens common between intents and snippets like method names and identifiers. This method achieves 22.3 BLEU on the CoNaLa dataset. We observe that it is outperformed by a standard LSTM seq2seq which scores around 24.34 BLEU when using variable standardization using an intent parser as proposed in~\cite{liguori2021evil,liguori2022generate}. A higher BLEU score using the standardization method is likely due to higher likelihood of generalization and more token overlap when standardizing tokens referring to variable names and values.

\textbf{Pretrained Transformer Models. } Mastropaolo et al.~\citeyearpar{mastropaolo2021studying} propose pretraining and fine-tuning a T5 model~\cite{raffel2019exploring} on a corpora of English text and source code and then fine-tune on several software engineering tasks including to generate assert statements (test methods) in Java. CodeBERT~\cite{feng_codebert_2020} is a pretrained transformer language model based on RoBERTa~\cite{liu2019roberta} (a variant of the BERT model~\cite{devlin2018bert}) pretrained on the CodeSearchNet corpus~\cite{husain_codesearchnet_2019}. CodeBERT has been used to generate exploit code snippets from natural language intents~\cite{liguori2021evil}. Orlanski and Gittens~\citeyearpar{orlanski2021reading} finetune the pretrained transformer BART~\cite{lewis-etal-2020-bart} on the CoNaLa dataset and achieve a performance of 26.24 BLEU which is comparable to that of CodeBERT scoring around 22.51 BLEU. Furthermore, BART when finetuned with both the CoNaLa annotated and mined datasets it achieved a BLEU score of 30.55. The authors also explore adding the question body along with the natural language intent as input to BART and find an increase in performance. BART achieves 34.35 when finetuned with question bodies and 35.32 when finetuned with questioned bodies on the both the annotated CoNaLa dataset and the mined CoNaLa datasets. Norouzi et al.~\citeyearpar{norouzi2021monolingual} find that a BERT-based encoder~\cite{devlin2018bert} and decoder model with a copying mechanism~\cite{gu2016incorporating} can achieve better performance by utilizing mine-able monolingual code data. The encoder is frozen on the monolingual code data and the decoder gets an additional objective during training of autoencoding the monolingual source code that corresponds to the intent that is fed into the encoder. This process is named Target Auto Encoding (TAE). Using the approach the decoder gets to also learn the source code meaning representation along with the encoder representation on the natural language intent. This approach achieves 32.57 BLEU on the CoNaLa dataset compared to the same set up without TAE which scores 30.98 BLEU. Scholak et al.~\citeyearpar{Scholak2021:PICARD} propose PICARD a simple and effective decoder-constraint algorithm that works with pre-trained encoder-decoder models. Using PICARD with a T5-3B model~\cite{raffel_exploring_2019} achieves state of the art on two SQL generation tasks from NL Spider~\cite{yuetal2018spider} and CoSQL~\cite{yu2019cosql}. GraphCodeBERT~\cite{Guo2020GraphCodeBERTPC} is a pretrained model for programming language that is pretrained using dataflow which encompasses the semantic structure of the code. The pretraining objectives used include masked language modeling, code structure edges, and representation alignment between source code and code structure. Other pretrained transformers used on source code include CodeT5~\cite{wang2021codet5}. CodeTrans~\cite{elnaggar2021codetrans}, PyMT5~\cite{clement2020pymt5}, CuBERT~\cite{kanade2020learning}, PLBART~\cite{ahmad-etal-2021-unified}, ProphetNet-X~\cite{qi2021prophetnet}, CoTexT~\cite{phan2021cotext}, T5-Code~\cite{mastropaolo2021studying}, GraphCodeBERT~\cite{Guo2020GraphCodeBERTPC}, and AlphaCode~\cite{alphacode}. Pretrained GPT-style Models for source code generation include CodeGPT~\cite{svyatkovskiy2020intellicode}, and GPT-Codex~\cite{chen2021evaluating}.

\begin{table*}[!htbp]

 \fontsize{8}{10}\selectfont

\renewcommand{\arraystretch}{1.5}

\begin{tabular}{p{3cm} p{6.2cm} p{3cm} p{3cm} }
\toprule


\multicolumn{1}{c}{\textbf{Area}}   & \multicolumn{1}{c}{\textbf{Description}} & \multicolumn{1}{c}{\textbf{Evaluation}}  & \multicolumn{1}{c}{\textbf{References}}   \\ \toprule 


Program Synthesis    & The task of synthesizing complete or partial programs given a specification, often structured input. & \% compilable programs, \% logically correct programs, synthesis duration, problems solved~\cite{schuster2021programming} \& other application specific metrics & \cite{shi2020tf, patra2016learning, gulwani2017program, cummins2017synthesizing, ellis2021dreamcoder,austin2021program}\\ \hline
Program Analysis  & Focuses on extracting semantic properties from programs and often classifying codebases using extracted properties.  & Accuracy, Precision, Recall, \& F1 & \cite{raychev2015predicting, mangal2015user,oh2015learning,panthaplackel2020associating,mou_convolutional_2016}\\ \hline
Program Repair   &  Focuses on building models to repair codebases given a compiler error message and a corresponding codebase. & full repair, partial repair, resolved error messages~\cite{hajipour2019samplefix}. & \cite{gupta2017deepfix,hajipour2019samplefix,Yasunaga20DrRepair,chirkova2020empirical, lutellier_coconut_2020, lu2021fapr,jiang_cure_2021}    \\ \hline
Clone Detection$^*$  &   Focuses on measuring the semantic similarity between  two  codes. There are variations of this task binary classification and retrieving code of semantic equivalence.  & Precision, Recall, F1, MAP~\cite{beitzel2009,CodeXGLUE} &   \cite{svajlenko2014towards,wang2020detecting, buch2019learning, wei2017supervised, chilowicz2009syntax} \\ \hline

Defect Detection$^*$  &   Focuses on detecting insecure code that can be exploited  in attacks such as DoS. The task is a binary classification  problem. & Accuracy, F1, Precision, Recall &   \cite{zhou2019devign, li2018vuldeepecker}   \\ \hline
Cloze Test$^*$   &   Focuses on predicting the correct token in a code function  given a natural language description and a code function  with a masked token. &  Accuracy &  \cite{husain_codesearchnet_2019,CodeXGLUE,feng_codebert_2020}    \\ \hline
Code Translation$^*$ &    Focuses on using machine learning to translate  source code  from a particular language (e.g. Python)  to another language (e.g. JavaScript). Also known as Code Transpiling. &  Exactness, BLEU~\cite{papineni2002bleu}, \& CodeBLEU~\cite{ren2020codebleu} &    \cite{lachaux2020unsupervised,feng_codebert_2020}    \\ \hline
Code Refinement$^*$   &   Focuses on automatically rewriting a codebase which is either buggy or convoluted. &   Exactness, BLEU~\cite{papineni2002bleu}, \& CodeBLEU~\cite{ren2020codebleu} &    \cite{hata2018learning,tufano2019empirical}   \\ \hline
Code Completion$^*$   &   Focuses on generating a code snippet that best completes a given program.& ROUGE~\cite{lin2004rouge}, exactness, edit similarity & \cite{svyatkovskiy2019pythia, svyatkovskiy2020intellicode,chirkova2020neural}      \\ \hline
Code Search and Retrieval$^*$ (Information Retrieval)  &   Retrieve a code snippet given a natural language query. & Mean Reciprocal Rank (MRR) \&  NDCG~\cite{vechtomova2009introduction,husain_codesearchnet_2019}  & \cite{feng_codebert_2020,husain_codesearchnet_2019, gu2018deepcodesearch}     \\ \hline
Code Documentation Generation$^*$  & Generating natural language descriptions of code snippets. Includes generating a function name for a function in code. Also known as Code summarization. & BLEU~\cite{papineni2002bleu} and human evaluation &  \cite{oda_learning_2015,feng_codebert_2020}      \\ \hline
Documentation Translation$^*$  &  Focuses on translating online code doc pages from one human language to another. & BLEU~\cite{papineni2002bleu} &    \cite{CodeXGLUE}   \\ \hline
 Code Generation$^*$    &   Generating code snippets given natural language intents. & Exactness, BLEU~\cite{papineni2002bleu}, semantic correctness~\cite{liguori2021evil}, pass@k~\cite{chen2021evaluating},  CodeBLEU~\cite{ren2020codebleu}  &  \cite{yin_tranx_2018,Xu2020IncorporatingEK,lin2018nl2bash}  \\ \hline
Semantic Parsing   &   Converting natural language to  unambiguous executable logical forms. & Exactness &  \cite{chen2019context,carpenter1997type, berant2014semantic}      \\  
\bottomrule
\footnotesize{$^*$ In CodeXGLUE}.

\end{tabular}

\caption{Overview of the areas in Code Intelligence (CI) , typical evaluation methods, and notable works in each area.}
\label{tab:nlp_code}


\end{table*}

\textbf{Transformer Models with Abstract Syntax Trees.} Dahal et al.~\citeyearpar{dahal-etal-2021-analysis} explore using tree representations of natural language utterances in code generation. Tree representations of natural language utterances are derived using constituency trees, which aim to describe the syntactic (grammatical) structure of an uttered sentence by dividing the sentence into sub-phrases. The authors run a series of experiments generating the AST representation of code using text-to-AST as a baseline, linearized tree-to-tree which uses the constituency tree of NL as input encoded using a standard transformer encoder, and structured tree-to-tree model. For the structured tree-to-tree model the authors utilize a structure aware Tree Transformer architecture~\cite{nguyen2020tree} with a pointer network~\cite{vinyals_2015_pointer} to copy tree leaves from the input to the output AST. The authors report improvements when using structure aware encodings of NL in their structured tree-to-tree model and achieve 30.30 BLEU on the CoNaLa dataset. Sun et al.~\citeyearpar{sun_treegen_2019} propose \textit{TreeGen} a neural architecture based on the Transformer model that incorporates AST grammar rules into the network. Incorporating ASTs into Transformers is observed to alleviate long-range dependencies between source code tokens~\cite{sun_treegen_2019,dahal-etal-2021-analysis} and better models input with multiple variables~\cite{dahal-etal-2021-analysis}.
\textbf{Multi-task Learning for Code Generation. }Multi-task learning in machine learning is training the model to perform multiple tasks simultaneously. Wei et al.~\citeyearpar{wei2020leveraging} propose using multi-task learning of code generation and summarization and report improvements in code summarization and retrieval benchmarks. Wei et al.~\citeyearpar{wei2019code} report improvements when dual training a model to generate code and to summarize code simultaneously. Parvez et al.~\citeyearpar{parvez2021retrieval} propose REDCODER, a code-search augmented model that retrieves code relevant to the inputted natural language description from an open source database to supplement the code generator model.



\subsection{Other Code Intelligence Areas}
Other areas of Code Intelligence (CI) include program repair, which focuses on building systems that automatically fix bugs in codebases using a compiler error message and a codebase as input. Notable program repair works include DeepFix~\cite{gupta2017deepfix} and DrRepair~\cite{Yasunaga20DrRepair}. Defect detection is another area of CI focused on detecting insecure code primarily through binary classification of codebases e.g.~\cite{li2018vuldeepecker,zhou2019devign}). Cloze Test is coding multiple choice problem where a model is tasked with predicting a masked code token. Code translation focuses on translating a codebase from one language (e.g. Java) to another programming language (e.g Python). There are various works in code translation~\cite{lachaux2020unsupervised,feng_codebert_2020}. Code refinement focuses on rewriting a codebase with the aim of simplification of removing bugs e.g.~\cite{hata2018learning,tufano2019empirical}. Lastly, code completion focuses on generating a code snippet that best completes a given codebase e.g.~\cite{svyatkovskiy2019pythia, svyatkovskiy2020intellicode,chirkova2020neural}. See Table~\ref{tab:nlp_code} for an overview of all the CI fields, metrics used in each field, and example works.


\subsection{Summary}
Application areas of NLP on source code such as program synthesis and program repair focus on automating redundant processes of software development. Program repair tends to focus on repairing programs given a compiler error message and is evaluated by percent of compilable programs, completely repaired (no logical errors) and partially repaired programs (some logical errors). Program synthesis tends to focus on generating full or partial programs often given structured input such as program examples, program description and output ~\cite{shi2020tf}. Synthesized programs are often evaluated using percentages (\%) of semantic and compilable programs. However, not all programming synthesis evaluate their programs, some works such as Cummins et al.~\citeyearpar{cummins_synthesizing_2017}, just focus on their application domain (program run-time speed-up) and discard generated programs that do not work. On the other hand, Program analysis is analogous to understanding language as opposed to generating it. It tends to focus on extracting features from developed code and using classification methods. CodeXGLUE lists several tasks of code intelligence and understanding. Those tasks are further listed along with all other known applications of NLP on source code in Table~\ref{tab:nlp_code}. The table also lists a description of the task, how the task is evaluated, and works that have participated in the task domain.

For an extensive survey on machine learning for big code c.f. Allamanis \emph{et al.} \citeyearpar{allamanis_survey_2018}, code generation and semantic parsing~\cite{lee2021toward}, and a short survey specifically on code generation from natural language~\cite{shin2021survey}. 

Code generation has promise in being integrated into conversational interfaces to enhance language understanding (e.g. ~\cite{artzi2011bootstrapping}). If code generation is integrated with conversational assistants we can see a rise of programming assistants which could be very helpful for programmers. Code generation may also be helpful for novice programmers and can have applications in fields such as computer science education. In the next section we will overview Conversational Artificial Intelligence (A.I.) and application of conversational artificial intelligence in education and software engineering.

\section{Conversational Artificial Intelligence}
\label{sec:conv_ai}

\begin{figure*}[h!]
  \includegraphics[width=\linewidth,height=3.5cm,keepaspectratio]{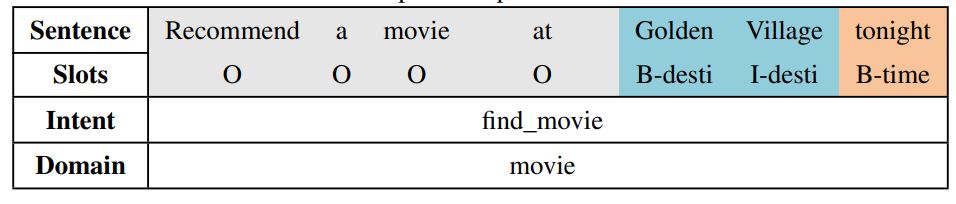}
  \caption{Example output from a Natural Language Understanding (NLU) module by~\cite{ni2021recent}. The utterance ``Recommend a movie at Golden Village tonight" is within the ``movie" domain and the user's goal is to ``find\_movie" at the destination ``Golden Village" with the time ``tonight". Here \texttt{desti} and \texttt{time} are slots and are accompanied by their corresponding slot-values ``Golden Village" and ``tonight" respectively.}
  \
  \label{fig:nlu}
\end{figure*}
In this section, we will focus on overviewing deep-learning-based automated conversational agents, also known as, dialogue systems. Conversational agents are popular and are widely accessible, from virtual sales agents to personal assistants like Google, Alex, or Siri, and have been applied in big domains such as general healthcare~\cite{montenegro2019health} and mental healthcare~\cite{weizenbaum1966eliza,d2017artificial}. There are generally two types of deep learning dialogue systems: Open-domain dialogue systems (also known as chit-chat chatbots) and task-oriented dialogue systems (also known as close-domain dialogue systems)~\cite{santhanam2019survey,ni2021recent}.
Task-oriented systems tend to be oriented towards a specific goal or a task such as booking a hotel, reserving a restaurant table, etc. Task-Oriented dialogue systems such as SimpleTOD~\cite{hosseini2020simple} are often entirely data-driven and are proficient in a certain set of domains given enough training examples within that domain. Task-oriented systems are often made up of 4 modules: Natural Language Understanding (NLU), Dialogue State Tracking (DST), Dialogue Policy Learning, and Natural Language Generation (NLG). NLU focuses on classifying what the user's goal is within a domain and parses out task-relevant words from the user utterance into \textit{slots} c.f. Figure~\ref{fig:nlu}. DST focuses on calibrating the dialogue state based on the current user utterance and the conversational history, it includes the user's goal and slot-value pairs. Dialogue Policy Learning is a module that decides the next action the system should take based on the dialogue state. NLG focuses on generating natural language from the selected dialogue action.  Open-domain dialogue systems such as MILABOT~\cite{serban_deep_2017} focus on chatting with a user without any domain restrictions and can cover a wide variety of topics~\cite{ram_conversational_2018}. For a survey on dialogue systems c.f.~\cite{santhanam2019survey,ni2021recent}. Next, we will overview works that apply task-oriented dialogue systems in software engineering and computer science education.

\subsection{Conversational Assistants for Software Engineering}



\begin{figure*}[h!]
  \includegraphics[width=\linewidth]{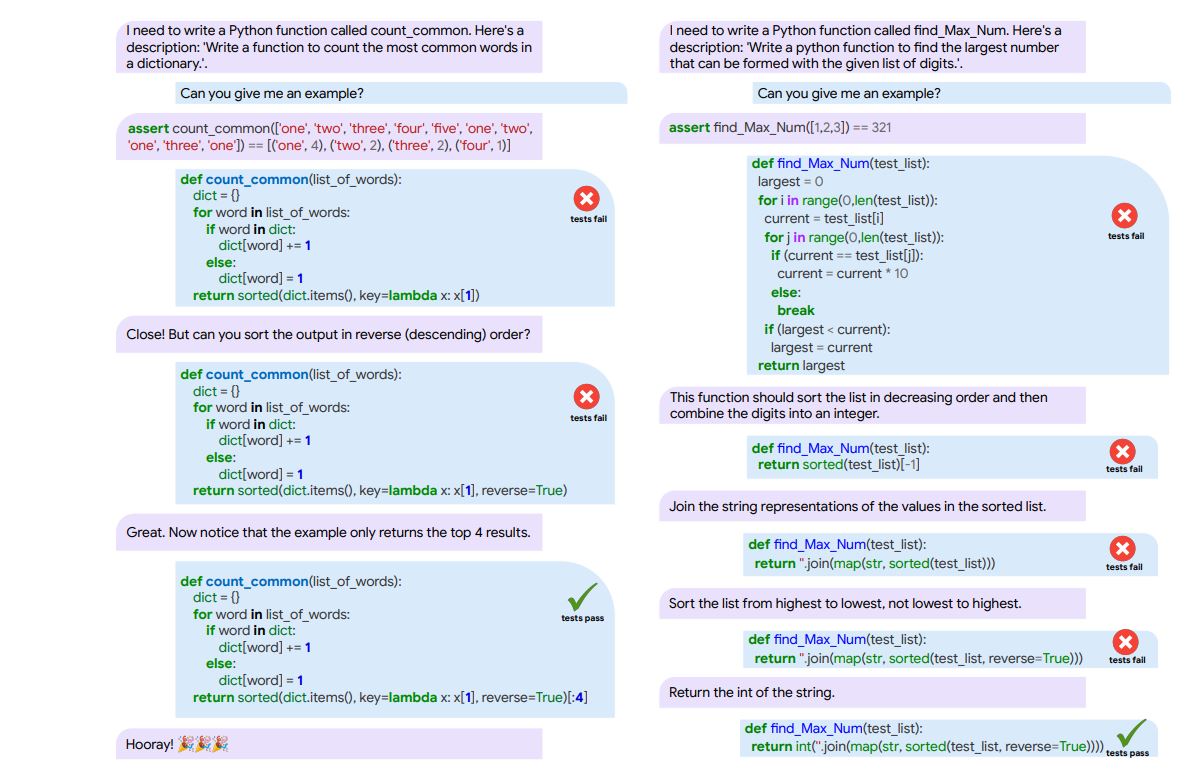}
  \caption{Example of human-AI conversational programming from~\cite{austin2021program}. Human utterances are shown in purple and AI utterances are denoted in blue.}
  \
  \label{fig:conv_prog}
\end{figure*}

This section focuses on overviewing some of the uses of conversational systems in the field of software engineering. We observe conversational assistants assisting programming in performing general programming tasks and assisting with specific software engineering workflows (e.g. Github version control actions).
\newline
\textbf{General Programming Assistants. }Chaurisa and Mooney~\citeyearpar{chaurasia2017dialog} propose a system that can engage a human user in dialogue for IFTTT code generation~\cite{quirk_language_2015}. However, the dialogue system mainly engages with the user to clarify their intent until the correct code is produced. Austin et al.~\citeyearpar{austin2021program} run controlled experiments with a simple dialogue system that generates Python code. The system mainly focuses on collaborating with a human to solve a particular programming task primarily through asking for clarifications and modifying the generated code as illustrated in Figure~\ref{fig:conv_prog}. These dialogue systems are simple and are mainly focused on solving one particular program, they neither include modules for dialogue state tracking nor code understanding. The dialogue systems also do not seek to collaborate with the human as much as get the human to help the system solve a particular programming task. The CoSQL dataset~\cite{yu2019cosql} does include some Task-Oriented dialogue components such as Dialogue State Tracking and user-system actions for dialogue policy learning, however, interfacing with a database via conversation does not assist with engineering new software. It is also worthy to mention IDE plugins and tools that utilize CI functionalities such as code search and code generation to assist software engineers and data scientists by generating or retrieving code using natural language specifications written by the user ~\cite{xu_ide_2021,heyman2021natural}. Xu et al.~\citeyearpar{xu_ide_2021} conduct a user study and find largely positive qualitative results however, the plugin's relationship with increased productivity, program quality, and correctness were inconclusive. 
\newline
\textbf{Conversational Assistants for Software Engineering Workflows. }Bradly et al.~\citeyearpar{bradley2018_context_aware} build \textit{Devy}, a conversational assistant with an NLU module that recognizes high-level Github-related intents and can execute a set of pre-defined low-level Github tasks. Paikari et al.~\citeyearpar{paikari2019chatbot} develop a task-oriented bot to inform software engineers of code conflicts using Github and to assist developers in locking/unlocking a source file for edits.


\subsection{Conversational Assistants for Computer Science Education}
Hobert~\citeyearpar{hobert_say_2019} proposes a coding tutoring system that is able to respond to open-ended knowledge questions, assess submitted source code, and guide students in natural language through a chat interface. The coding tutor's interface has a coding area and a chat area where the user is able to converse with the automated coding tutor. The task-oriented dialogue tutor contains intent and entity recognition modules and a natural language generation module. It also has a set of predefined ``learning paths" based on a Finite State Machine stored in a database. Responses to student questions and guidance given by the system are all derived from the predefined finite state machine specific to a programming problem. To evaluate the coding tutor, a questionnaire was handed out to students asking them to rate the system on usefulness, ease of use, practice, and whether students would use the system again.

\section{Future Directions}
\label{sec:future}



This section explores possible future directions of research at the intersection between conversational systems and CI. We focus on the two areas of computer science education and software engineering.

\subsection{Programming-Oriented Dialogue Systems} 
Task-oriented dialogues currently focus on assisting the human user in completing API-centered redundant tasks such as booking a hotel. Current advances in Code Intelligence research such as the availability of big data containing pairs in both code and natural language, code search, program repair, code generation, and recently program execution tracing~\cite{nye2021show}, enable us to create more intelligence tools to enable professional and novice programmers. There is room to create dialogue systems that can assist humans in specific programming tasks such as code-refactoring, generating methods from NL descriptions, retrieving code examples, bug fixing, and explaining portions of the code in natural language through conversation. Some of these tasks are currently automated through various IDE plugins e.g.~\cite{xu_ide_2021} however there is a back-and-forth process between a system and a programmer that can be captured well in dialogue. To support such capabilities, task-oriented dialogue systems can be adapted. The natural language understanding module in task-oriented dialogue systems can be enhanced with a code-understanding module, and likewise, with the natural language generation component, it can be coupled with a code generation module. To the best of our knowledge, there exist no programming-oriented dialogue datasets which are crucial to driving this research area forward.

\subsection{Computer Science Education} 
\textbf{Self-Regulated Learning (SRL). } SRL is a framework for understanding how students control their behavior during learning and it includes: 1) Cognitive processes related to content knowledge, reasoning, and problem-solving; 2) Metacognitive processes where the learner plans their learning endeavors, and in identifying gaps in knowledge or seeking help; 3) Affective processes that include a student's goals and emotional states.~\cite{pintrich2000slr}. There is potential in adapting dialogue systems to deliver personalized learning interventions through a dialogue system. Example cognitive interventions that can be delivered through a conversational assistant with CI include: Generating code-related hints to students in the form of Socratic questions, also known as guided-inquiry i.e., smaller subquestions that can guide to a final answer. Cobbe et al.~\citeyearpar{cobbe2021training} utilize GPT-3~\cite{brown2020language} and finetune on ~800 examples to generate mathematical Socratic subquestions by conditioning each subquestion on a ground-truth mathematical step in the solution.\footnote{See an example here: \href{https://github.com/openai/grade-school-math}{https://github.com/openai/grade-school-math}}. Other interventions can include generating example code as hints, code repair assistance, and recommending course content. Metacognitive interventions by code-aware dialogue systems can include identifying knowledge gaps from coding patterns and mistakes and suggesting that the student asks the human Teaching Assistant (TA) for help. 





\textbf{IDE-based Learning Analytics. }IDE-based learning analytics utilizes Integrated Development Environments (IDEs) to collect data about learners' programming patterns and deliver learning interventions \cite{hundhausen_ide-based_2017}. Programming data collected from IDEs can include: (1) editing data (e.g., code snapshots), (2) compilation data (e.g., compilation errors), (3) execution data (e.g., run-time exceptions), and (4) debugging data (e.g., breakpoints, steps, and inspecting variables). Hundhausen et al.~\citeyearpar{hundhausen_ide-based_2017} proposed a four-phase process model for IDE-based data analytics consisting of: (1) data collection, (2) data analysis, (3) intervention design, and (4) intervention delivery. There is extensive work in data collection and data analysis in IDE-based learning analytics~\cite{watson_predicting_2013,diana_measuring_2018,ahadi_students_2016,carter_normalized_2015} however to the best of our knowledge, there is little work on delivering automated interventions. Interventions are primarily focused on fixing syntax errors~\cite{bhatia2016automated}, enhanced error messages~\cite{becker_effective_2016}, and generating hints to programmers~\cite{chow_automated_2017,rivers2017data}. Carter et al~\citeyearpar{carter_normalized_2015} proposes the Programming State Model (PSM), which ``categorizes students' programming within a two-dimensional space that captures both a student's current activity (e.g., editing, debugging) and the correctness of the student's most recently compiled programming solution". A dialogue system with PLP capabilities can automate the process of intervention design and delivery when integrated with IDE-analytics. Furthermore, it can work with an IDE-based learning analytics code understanding module. A programming-oriented dialogue system in which the user intents are derived from learning analytics-inspired IDE-activity similar to the proposed Programming State Model. The dialogue policy learner would decide what intervention to perform given a set of pre-defined educational interventions. Dorodchi et al.~\citeyearpar{dorodchi_custom_2020} propose a custom IDE prototype that can be integrated with a dialogue system to deliver personalized learning interventions.

\subsection{Human Computer Interaction} Another area of research can focus on designing programmer-centered IDE interfaces around \textit{conversational programming}, where both a human and a conversational agent with CI can collaborate on programming. A glimpse of conversational programming can be seen in Figure~\ref{fig:conv_prog} from Austin et al.~\citeyearpar{austin2021program}.

\section{Conclusion}
\label{sec:conclusion}

In this paper, we overviewed the field of code intelligence (CI), which focuses on applying artificial intelligence (AI) to source code. We identified 14 key areas of research in CI summarized in Table~\ref{tab:nlp_code}. Through analyzing existing works in code generation using the CoNaLa dataset we observe a trend of Transformer-based models being more heavily used in recent years and pushing state-of-the-art boundaries on this task see Figure~\ref{fig:conala}. We summarize our findings from systematically reviewing works that propose and curate datasets containing natural language and source code ( Table~\ref{tab:code_datasets}). Conversational AI is generally divided into two areas of research: Open-domain dialogue systems which can chat about a wide variety of topics, and task-oriented dialogue systems which focus on assisting the user with completing a specific goal or a task. We identify existing works in task-oriented dialogue systems with CI capabilities, mainly falling into assisting professional programmers with the software engineering process or assisting novice programmers in educational contexts. We identify future directions for research at the intersection of dialogue systems and code intelligence. The first major direction is Programming-Oriented Dialogue Systems (PODS), which are task-oriented dialogue systems with CI capabilities.

We identify a dire need for dialogue datasets for this area. Dialogue datasets can then be utilized to create PODS to assist professional developers by having a dialogue system encompass many existing code intelligence plugins. We also identify a research opportunity for POD systems to assist computer science learners by facilitating the self-regulated learning process. Enhancing POD systems with learning analytics would enable PODS to learn more about the student user and intervene more appropriately. Finally, we identify research opportunities in human computer interaction to create custom interfaces where a PODS and a user can collaborate on solving programming problems effectively.





\bibliography{paper.bib}
\bibliographystyle{acl_natbib}

\end{document}